\documentclass[lettersize,journal]{IEEEtran}
\usepackage{amsmath,amsfonts}
\usepackage{algorithmic}
\usepackage{array}
\usepackage{textcomp}
\usepackage[font=small]{caption}
\usepackage{enumitem}
\usepackage{subfigure}
\usepackage{url}
\usepackage{verbatim}
\usepackage{graphicx}
\usepackage[printonlyused]{acronym}
\usepackage[maxbibnames=10,style=ieee]{biblatex}
\usepackage{makecell}
\usepackage{color}
\usepackage{xcolor}
\usepackage{fancyhdr}

\colorlet{blue}{black}
\usepackage{balance}

\fancypagestyle{firststyle}
{
   \fancyhf{}
   \fancyhead[C]{\footnotesize{This article has been accepted for publication in IEEE Transactions on Affective Computing. This is the author's version which has not been fully edited and
content may change prior to final publication. Citation information: DOI 10.1109/TAFFC.2026.3679039.}}
}

\begin{document}
\title{Towards Emotion Recognition with 3D Pointclouds Obtained from Facial Expression Images}
\vspace{4mm}
\author{Laura Rayón Ropero\IEEEauthorrefmark{1}, Jasper De Laet\IEEEauthorrefmark{1}, Filip Lemic\IEEEauthorrefmark{2}, Pau Sabater Nácher, Nabeel Nisar Bhat, Sergi Abadal, Jeroen Famaey, Eduard Alarcón, Xavier Costa-Pérez\vspace{-2mm}
\IEEEcompsocitemizethanks{\IEEEcompsocthanksitem\IEEEauthorrefmark{1}Equal contribution.}
\IEEEcompsocitemizethanks{\IEEEcompsocthanksitem\IEEEauthorrefmark{2}Corresponding author.}
\thanks{L. Rayón, E. Alarcón, and S. Abadal are affiliated with the Universitat Politècnica de Catalunya, Spain, email: \{name.surname\}@upc.edu.}%
\thanks{J. De Laet, N. N. Bhat, and J. Famaey are affiliated with the University of Antwerp, Belgium, email: \{name.surname\}@uantwerpen.be. N. N. Bhat and J. Famaey are additionally affiliated with imec, Belgium.}
\thanks{F. Lemic, P. Sabater, and X. Costa-Pérez are affiliated with i2CAT Foundation, Spain, email: \{name.surname\}@i2cat.net. F. Lemic is also with the Faculty of Electrical Engineering and Computing, University of Zagreb, Croatia. X. Costa-Pérez is also affiliated with the NEC Labs Europe GmbH, Germany and ICREA, Spain.
}
\thanks{This work has been supported by the Smart Networks and Services Joint Undertaking (SNS JU) under the European Union's Horizon Europe research and innovation programme (grants nº 101139161 - INSTINCT and nº 101192521 - MultiX projects). The work of Nabeel Nisar Bhat was supported by the Fund for Scientific Research Flanders (FWO), grant nº 1SH5X24N.
}}

\markboth{IEEE DRAFT}%
{IEEE DRAFT}

\maketitle
\thispagestyle{firststyle}

\begin{abstract}
\ac{FER} is a critical research area within \ac{AC} due to its wide-ranging applications in \ac{HCI}, as well as its potential use in mental health assessment and fatigue monitoring. However, current FER methods predominantly rely on \ac{DL} techniques trained on \ac{2D} image data, which pose significant privacy concerns and are unsuitable for continuous, real-time monitoring. As an alternative, we propose \ac{JCAS} as an enabler of continuous, \textcolor{blue}{privacy-aware} \ac{FER}, through the generation of detailed \ac{3D} facial pointclouds via on-person sensors embedded in wearables. We present arguments supporting the privacy advantages of \ac{JCAS} over traditional \ac{2D} imaging approaches, particularly under increasingly stringent data protection regulations. A major barrier to adopting \ac{JCAS} for \ac{FER} is the scarcity of labeled \ac{3D} \ac{FER} datasets. Towards addressing this issue, we introduce a method based on the \ac{FLAME} model to generate \ac{3D} facial pointclouds from existing public \ac{2D} datasets. Using this approach, we create AffectNet3D, a \ac{3D} version of the AffectNet database. To evaluate the quality and usability of the generated data, we design a pointcloud refinement pipeline focused on isolating the facial region, and train the popular PointNet++ model on the refined pointclouds. Fine-tuning the model on a small subset of the unseen \ac{3D} \ac{FER} dataset BU-3DFE yields a classification accuracy exceeding 70\%, comparable to oracle-level performance. To further investigate the potential of \ac{JCAS}-based \ac{FER} for continuous monitoring, we simulate wearable sensing conditions by masking portions of the generated pointclouds. Experimental results show that models trained on AffectNet3D and fine-tuned with just 25\% of BU-3DFE significantly outperform those trained solely on BU-3DFE. These findings highlight the viability of our data generation pipeline and support the feasibility of continuous, \textcolor{blue}{privacy-aware} \ac{FER} via wearable \ac{JCAS} systems.
\end{abstract}



\acrodef{FER}{Facial expression-based Emotion Recognition}
\acrodef{AC}{Affective Computing}
\acrodef{HRI}{Human-Robot Interaction}
\acrodef{JCAS}[HFWS]{High-Frequency Wireless Sensing}
\acrodef{TL}{Transfer Learning}
\acrodef{I2P-MAE}{Image-to-Point Masked Autoencoders}
\acrodef{BU-3DFE}{Binghamton University 3D Facial Expression}
\acrodef{2D}{2-Dimensional}
\acrodef{3D}{3-Dimensional}
\acrodef{EU}{European Union}
\acrodef{DL}{Deep Learning}
\acrodef{ML}{Machine Learning}
\acrodef{CAD}{Computer-Aided Design}
\acrodef{CNN}{Convolutional Neural Network}
\acrodef{MIMO}{Multiple Input Multiple Output}
\acrodef{RGB}{Red, Green, Blue}
\acrodef{THz}{Terahertz}
\acrodef{HCI}{Human Computer Interaction}
\acrodef{FACS}{Facial Action Coding System}
\acrodef{AU}{Action Unit}
\acrodef{FL}{Facial Landmark}
\acrodef{MLP}{Multi-Layer Perceptron}
\acrodef{PCA}{Principal Component Analysis}
\acrodef{RAF-DB}{Real-world Affective Faces Database}
\acrodef{DDAMFN}{Dual-Direction Attention Mixed Feature Network for FER}
\acrodef{FLAME}{Faces Learned with an Articulated Model and Expressions}
\acrodef{SotA}{State-of-the-Art}
\acrodef{FMCW}{Frequency-modulated Continuous Wave}
\acrodef{SAR}{Synthetic Aperture Radar}
\acrodef{MCS}{Modulation and Coding Scheme}
\acrodef{EM}{Electromagnetic}
\acrodef{CW}{Continuous Wave}
\acrodef{AI}{Artificial Intelligence}
\acrodef{HMD}{Head Mounted Device}
\acrodef{xR}{Extended Reality}
\acrodef{FPS}{Farthest Point Sampling}
\acrodef{SfM}{Structure-from-Motion}
\acrodef{GDPR}{General Data Protection Regulation}
\acrodef{PT}{Point Transformer}
\acrodef{DGCNN}{Dynamic Graph CNN}
\acrodef{GCN}{Graph Convolutional Network}
\acrodef{SoC}{System on a Chip}
\acrodefplural{SoC}{Systems on a Chip}
\acrodef{FLOP}{Floating-Point Operations per Second}
\acrodef{CUDA}{Compute Unified Device Architecture}
\acrodef{CPU}{Central Processing Unit}
\acrodef{GPU}{Graphics Processing Unit}
\acrodef{NPU}{Neural Processing Unit}
\acrodef{RAM}{Random Access Memory}

\vspace{-3mm}
\section{Introduction}

The exploration of computational models for recognizing and interpreting human emotions has been an active research area for over three decades~\cite{ojha2021computational}. This interest stems from humanity's curiosity about its psyche and the ambition to improve quality of life through emotional well-being. Emotion-aware computational models fall under the umbrella of \acf{AC}~\cite{zhang2020emotion}, which leverages computer science, psychology, and cognitive science, to create systems that can recognize human emotions, thus enhancing \acf{HCI}. Facial expressions, generated by facial muscle movements, are key non-verbal communication cues~\cite{hall2022nonverbal}. Within \ac{AC}, the concept of continuous emotional monitoring has emerged as a promising yet underexplored direction, particularly in domains such as healthcare and education. Research in these fields highlights the importance of long-term affective observation, with studies such as~\cite{loveland1997emotion,yirmiya1989facial,esposito2011assessment} linking mental diseases and neurodivergence to traceable emotional patterns, detectable only through day-to-day tracking.

The \acf{2D} \acf{FER}, also referred to as the image-based \ac{FER}, was one of the first FER methodologies explored within \ac{AC}. The early models were based on handcrafted features and geometric facial models \cite{ko2018brief} \cite{massoli2021mafer}. 
A significant limitation of these methods was that they needed to be tailored for specific datasets or targets, resulting in low re-usability \cite{vignesh2023novel}. The proliferation of large FER-labeled databases such as \ac{RAF-DB}~\cite{shan2018reliable} (containing around 30K facial images), and AffectNet~\cite{mollahosseini2017affectnet} (containing around 400K facial images), enabled a transition to \ac{DL} architectures. These models require a high amounts of training data to prevent overfitting, but have also demonstrated improved abstraction and generalization abilities \cite{canal2022survey}.

\begin{figure*}
\centering
\begin{minipage}{.33\textwidth}
  \centering
  \includegraphics[width=0.99\linewidth]{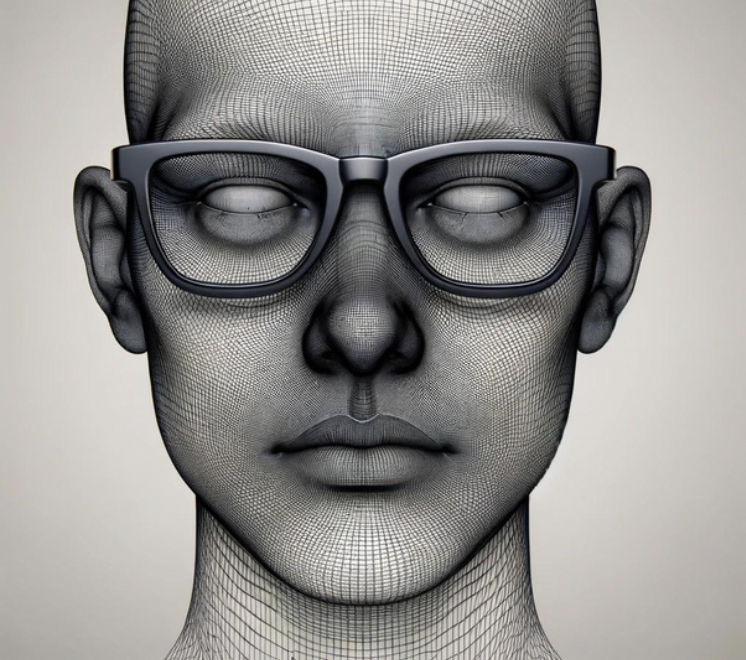}
  \vspace{-2mm}
  \captionof{figure}{\textcolor{blue}{Privacy-aware} 3D FER through \acf{JCAS}}
  \label{fig:jcas_paradigm}
  \vspace{-1mm}
\end{minipage}\hfil
\begin{minipage}{.59\textwidth}
  \centering
  \includegraphics[width=0.93\linewidth]{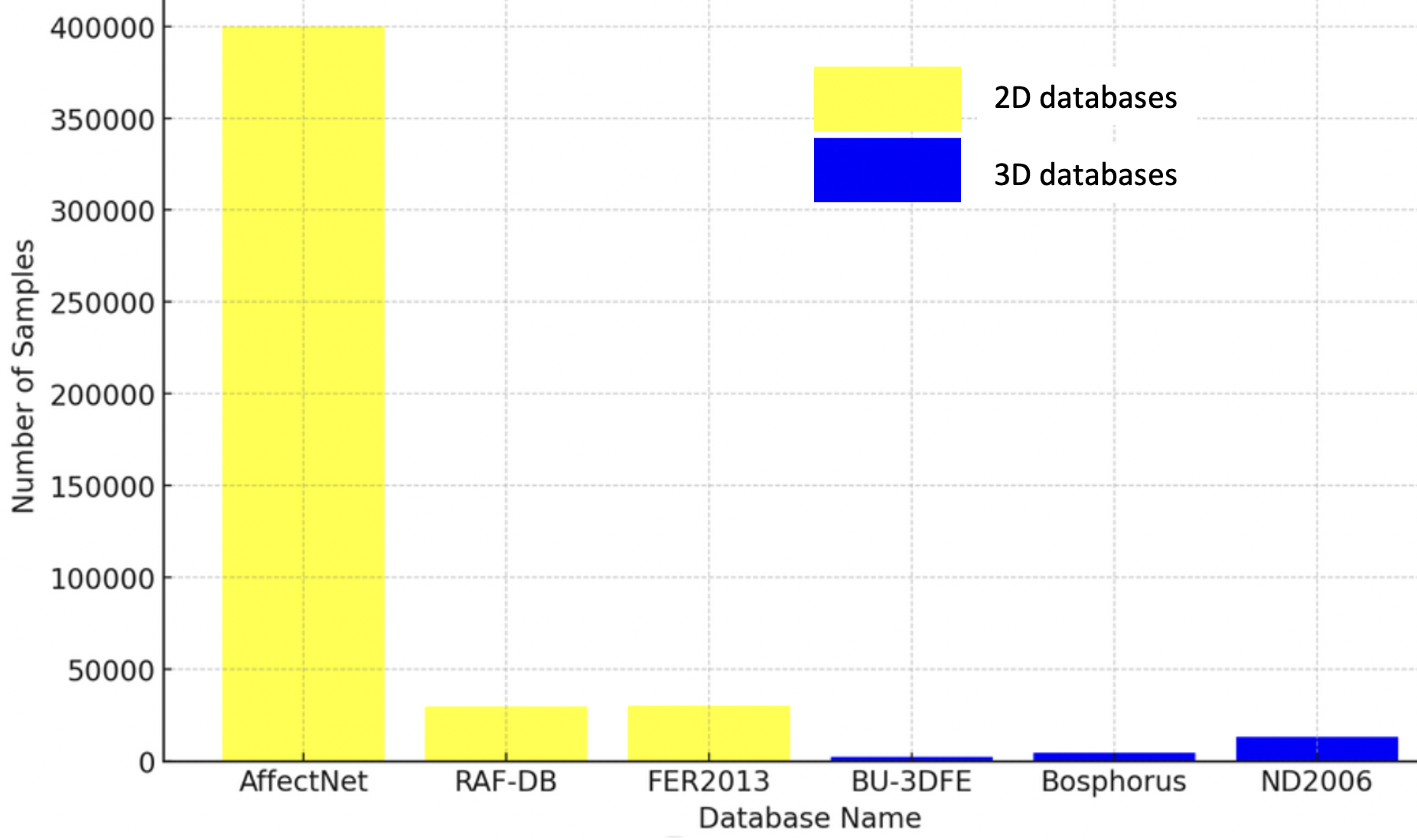}
   \vspace{-1mm}
  \captionof{figure}{Comparison between 2D and 3D FER databases}
  \label{fig:2d_vs_3d}
\end{minipage}
\vspace{-4mm}
\end{figure*}

While image-based \ac{FER} methods have achieved impressive results, they face significant challenges:

\begin{enumerate}[leftmargin=*]
    \item The acquisition of \ac{2D} images introduces privacy concerns under global regulations such as the EU \ac{AI} Act and \ac{GDPR}~\cite{butt2024analytical,morandin2023ten}.
    \item In scenarios requiring continuous facial monitoring, traditional image acquisition systems (e.g., cameras) are often impractical, intrusive, and uncomfortable for long-term use.
\end{enumerate}

These limitations have motivated research into alternative modalities such as the \ac{3D} pointcloud-based \ac{FER}, which enhances privacy by removing high-resolution facial textures while preserving essential geometric features. However, \ac{3D} \ac{FER} introduces its own set of challenges, particularly for training \acf{DL} models. The acquisition of \ac{3D} facial data is inherently complex and costly, often requiring sophisticated equipment such as multi-camera \ac{3D} scanners~\cite{yin20063d}. As a result, existing \ac{3D} \ac{FER} datasets (such as \ac{BU-3DFE}~\cite{yin20063d} and Bosphorus~\cite{savran2008bosphorus}) are significantly smaller than their \ac{2D} counterparts. This scarcity increases the risk of overfitting and limits the generalization of \ac{DL}-based models to real-world conditions. Furthermore, most available \ac{3D} datasets contain posed expressions captured in controlled environments, whereas \ac{2D} datasets such as AffectNet~\cite{mollahosseini2017affectnet} and \ac{RAF-DB}~\cite{shan2018reliable} include large-scale, diverse, and \textit{in-the-wild} samples. The absence of an equivalent \textit{in-the-wild} \ac{3D} \ac{FER} dataset remains a critical bottleneck, largely due to the difficulty of acquiring \ac{3D} data outside studio conditions.

A promising avenue to overcome these limitations lies in leveraging \acf{JCAS} (cf., Figure~\ref{fig:jcas_paradigm}) for facial imaging~\cite{wu2022sensing}. \ac{JCAS} repurposes communication signals for sensing, eliminating the need for dedicated sensors, thereby reducing hardware complexity, energy consumption, and privacy risks. Wearables equipped with \ac{JCAS}, such as smart glasses, can continuously generate \ac{3D} facial maps, enabling non-intrusive and \textcolor{blue}{privacy-aware} continuous \ac{FER}~\cite{hong201860,zubair2024high,singh2020design}. \textcolor{blue}{In this paper, we use the term \emph{privacy-aware} to denote a representation-level reduction of directly identifiable visual cues, achieved by relying on texture-free 3D geometry rather than high-resolution 2D imagery. We do not claim formal privacy guarantees against re-identification, inversion, or reconstruction attacks.}
In remote intelligent health systems, this capability supports affect-aware and privacy-compliant monitoring, aligning with applications such as tele-consultations and fatigue tracking. By combining \ac{JCAS} 3D \ac{FER} with physiological data, such as heart rate or respiration~\cite{petkie2008remote}, comprehensive user profiles can be obtained without increasing the intrusiveness of the sensing infrastructure.

For \ac{JCAS}-enabled systems to perform emotion recognition effectively, they require robust \ac{DL} models trained on diverse \ac{3D} data accurately reflecting real-world conditions. Since data captured by \ac{JCAS} will inherently be \textit{in-the-wild}, the lack of large-scale \ac{3D} \ac{FER} datasets presents a fundamental challenge (cf., Figure~\ref{fig:2d_vs_3d}). To address this, we explore the feasibility of generating extensive labeled \ac{3D} \ac{FER} datasets from existing labeled \ac{2D} databases using statistical face reconstruction techniques. By converting \ac{2D} images into \ac{3D} pointclouds, our approach leverages the wealth of available \ac{2D} \ac{FER} data to train models capable of generalizing to real-world \ac{3D} \ac{FER} scenarios, including those enabled by \ac{JCAS}-based sensing.
By relying on \ac{3D} pointclouds without high-resolution textures, the framework presented in this study mitigates identity disclosure risks (a major barrier to adopting \ac{FER} in healthcare environments), while ensuring compatibility with \ac{JCAS} enabled wearables for continuous, regulation-compliant, and \textcolor{blue}{privacy-aware} emotion recognition.

\subsection{Contributions}

This work makes the following contributions:

\begin{itemize}[leftmargin=*]

    \item We investigate the feasibility of using \ac{2D} image-generated pointclouds as a training dataset for \ac{3D} \ac{FER}, leveraging publicly available in-the-wild \ac{2D} \ac{FER} datasets to overcome the scarcity of labeled \ac{3D} \ac{FER} data. This approach significantly reduces data collection burdens while improving model generalization. Using a contemporary \acf{FLAME}-based \ac{3D} reconstruction techniques~\cite{li2017learning}, we convert AffectNet \ac{2D} images into \ac{3D} pointclouds and train a \ac{3D} \ac{DL} model.

    \item We explore the use of \ac{3D} facial imaging for real-world inference, showing that publicly available \ac{2D} \ac{FER} datasets can be repurposed for \textcolor{blue}{privacy-aware} \ac{3D} \ac{FER}. To enhance robustness, we develop a data refining pipeline that isolates facial regions in \ac{3D} pointclouds, ensuring models trained on \ac{JCAS}-like data maintain high classification accuracy. To assess the feasibility of \ac{JCAS}-based \ac{FER} for continuous monitoring, we simulate wearable sensing by selectively masking portions of the generated pointclouds, replicating occlusion constraints.

    \item We fine-tune our trained \ac{3D} \ac{FER} model on a small subset of the unseen \ac{3D} \ac{FER} dataset (\ac{BU-3DFE}) and evaluate its generalization performance. 
    Our results show that this approach achieves classification accuracy comparable to an oracle solution trained on the entire \ac{3D} dataset, significantly outperforming baselines trained solely on small \ac{3D} datasets.
\end{itemize}

\subsection{Paper Structure}

This paper is structured as follows. 
Section~\ref{sec:background} provides the necessary background on \ac{FER} and the role of \ac{3D} pointclouds in emotion recognition. 
Section~\ref{sec:related_works} reviews prior works in \ac{DL} architectures for face reconstruction from images, providing a rationale for selecting a \ac{FLAME}-based method for AffectNet \ac{3D} generation, as well as for the utilizing PointNet++ for 3D FER. 
In addition, the section outlines literature efforts outlining advances in THz-based 3D reconstruction, paving the way toward short-range and small-scale \ac{JCAS} systems for \ac{FER}. 
Sections~\ref{sec:methodology} and~\ref{sec:results} present our experimental methodology and evaluation results for the proposed \ac{3D} \ac{FER} model, respectively. 
Finally, Section~\ref{sec:discussion} discusses the broader implications of our findings and potential future research directions, while Section~\ref{sec:conclusion} concludes the paper.


\section{Background}
\label{sec:background}

\subsection{Theoretical Foundations of FER}

Theoretical foundations of \ac{FER} technology have their roots in the Charles Darwin’s Expression of the Emotions in Man and Animals~\cite{newmark2022charles}. 
Darwin treated emotions as separate discrete entities and, presumably driven by his global travels aboard the HMS Beagle, stated that facial expressions of emotion are universal.
These foundational ideas led subsequent emotion theorists to the development of several models about affective facial behaviors, i.e., how emotions are represented and recognized through facial expressions. 
The most utilized among these models include the Categorical, Dimensional, and \ac{FACS} models~\cite{mollahosseini2017affectnet}.

The Categorical model links facial expressions to a discrete set of basic and universal emotion categories (e.g., happiness, surprise, anger)~\cite{ekman1971constants}. 
The compound version emerged as a way to handle samples displaying a mixture of two emotions, by defining a set of compound emotion categories that are derived from the Categorical model (i.e., happily surprised, sadly fearful)~\cite{du2014compound}. 
The Dimensional model appeared as an alternative to the Categorical model, proposing a continuous rather than discrete representation of emotions. 
A notable implementation of this model is the Valence-Arousal scale, where Valence assesses the positivity or negativity of an emotion, and Arousal measures its level of excitement or calmness~\cite{russell1980circumplex}. 
Differently from these models, the \ac{FACS} takes an anatomical approach, focusing not on inferring affective states directly, but on identifying changes in facial muscles. 
This model codes the actions of individual or groups of muscles present in facial expressions~\cite{ekman1978facial}. 

\subsection{Technological Foundations of FER}

The Categorical model is the most prevalent in the Computer Vision community. 
Consequently, it serves as the basis for labeling most \ac{FER} databases. 
This model, proposed by discrete emotion theorist Paul Ekman~\emph{et al.}~\cite{ekman1971constants}, initially categorized facial expressions into six basic emotions: anger, disgust, fear, happiness, sadness, and surprise. 
Subsequent research added contempt~\cite{xie2022overview} and  neutral emotions to the list. 
Both databases employed within this work, the 3D version of AffectNet and the BU-3DFE database, are labelled according to Ekman's 6-emotion categorical model. 
They each include the ``neutral'' category, while AffectNet additionally incorporates the ``contempt'' category.

\ac{FER} typically involves 3 steps: 
i) pre-processing consisting of applying transformations to the data samples for facilitating feature extraction;
ii) feature extraction that aims to extract from the data samples the facial features that convey emotion;  and
iii) classification according to an emotion model.

FER studies can be broadly categorized based on the nature of the data they use as static and dynamic FER~\cite{ko2018brief, dujaili2024survey}. 
Static FER relies on static facial features. 
In contrast, dynamic FER, also known as video-based FER, uses spatio-temporal features to capture the dynamics in facial expression sequences. 
Typically, dynamic FER delivers higher accuracy rates than static FER due to its incorporation of temporal data.
However, dynamic FER faces significant challenges; dynamic features can vary significantly in their duration and characteristics based on the individual. 
Furthermore, the process of temporal normalization, which aims to standardize expression sequences to a consistent frame count, often results in the loss of essential temporal details that are critical for accurate emotion recognition. 
Due to these complexities, and the fact that static FER databases are more numerous, research in static FER has been more prevalent and it is in the  focus of this study. 
However, considering the anticipated energy-efficient characteristics of the devices developed within the \ac{JCAS} paradigm, high sampling frequencies could enable the exploration of dynamic FER.

\subsection{\ac{2D} vs. \ac{3D} FER}

FER databases can be categorized by the dimension of their data samples as \ac{2D} or \ac{3D}. 
\ac{2D} samples consist of labeled images that capture facial expressions in a flat plane. 
3D samples encompass models that represent facial expressions in a volumetric space. 
This work bridges the 2D and 3D worlds by exploring whether 3D pointclouds inferred from 2D images can be used for DL model training for 3D-FER tasks.

2D FER was among the earliest explored FER methodologies, with the early models primarily based on handcrafted features and geometric facial priors~\cite{ko2018brief,massoli2021mafer}, some of which continue to be used today, sometimes in conjunction with \ac{DL} architectures~\cite{pham2021facial}. 
This continued usage is attributed to their ability for solving typical FER challenges such as variations in illumination, pose, and occlusion. Nonetheless, a significant limitation of these methods is that they need to be tailored for specific datasets or targets, resulting in low reusability~\cite{vignesh2023novel}.

A major contribution of \ac{DL} to 2D \ac{FER} has been the introduction of \acp{CNN}, which automate the process of feature extraction and have become the foundational components of modern FER architectures. \acp{CNN}, together with ensemble methods that combine multiple network outputs to enhance robustness and predictive power~\cite{pham2021facial}, have improved the reliability of 2D FER systems. Current benchmarks on widely used 2D FER datasets, such as RAF-DB and AffectNet, reveal that leading architectures typically rely on either CNN layers~\cite{vignesh2023novel, pham2021facial, zhang2023dual, le2024patt} or Transformer blocks~\cite{mao2023poster++, wasi2023arbex} as their primary building units.

The main difficulty in FER lies in the phenomenon of small inter-class differences coupled with large intra-class variability. Emotion categories often exhibit subtle distinctions from one another (inter-class similarity), while expressions of the same emotion can vary widely across individuals (intra-class variability). To address this, recent models have incorporated attention mechanisms and multi-scale learning strategies as core design principles. Attention mechanisms enable networks to focus on key facial regions (such as the eyes, eyebrows, nose, and mouth) thereby improving sensitivity to emotionally relevant features while suppressing redundant or irrelevant information. Multi-scale learning, in turn, allows models to integrate both fine-grained and global facial features, improving generalization and robustness.

CNN-based methods often implement attention mechanisms through U-Net-style segmentation~\cite{vignesh2023novel, pham2021facial} or masking strategies~\cite{pham2021facial}, and employ residual connections to facilitate multi-scale feature learning~\cite{pham2021facial, vignesh2023novel, zhang2023dual}. Other variants introduce dual attention heads~\cite{zhang2023dual} or transformer-inspired dot-product self-attention modules~\cite{le2024patt} to better capture spatial dependencies. Transformer based methods include the use of Visual Transformers (ViT) \cite{wasi2023arbex}, cross-fusion transformer encoders \cite{mao2023poster++}, and window-based cross-attention \cite{mao2023poster++,wasi2023arbex}. Multi-scale learning in transformer-based methods is tackled through pyramidal \cite{mao2023poster++} or multi-level \cite{wasi2023arbex} architectures.

These key architectural characteristics are not exclusive to 2D-based FER; they are widely shared with emerging 3D-based FER approaches. This convergence underscores a fundamental principle: the essence of facial expressions transcends the dimensionality of the data.

3D-based FER is a relatively novel field in Affective Computing. Whereas 2D-based FER was already well-established by the early 2000s, 3D FER began gaining attention later as a means to overcome challenges presented by 2D images, such as variations in pose and illumination ~\cite{ko2018brief}. In image-based settings, lighting conditions and changes in the subject's pose can drastically affect the visibility and appearance of facial features, complicating the process of emotion recognition. With the emergence of the \ac{JCAS} paradigm, the exploration of 3D-based FER has become even more critical.

Nonetheless, 3D FER introduces its own challenges, particularly in the training of \ac{DL} models. While 2D FER already benefits from large, readily available databases compiled from internet-sourced data~\cite{mollahosseini2017affectnet,shan2018reliable,goodfellow2013challenges}, producing equivalent 3D databases is more challenging. 
3D data capture still requires controlled studio settings and involves more expensive and sophisticated equipment, such as 3D scanners composed of several cameras~\cite{yin20063d,zhang2013high}. 
This results in a limited availability of 3D facial expression data.
In addition, occlusions such as hair, glasses, or other objects that partially cover the face present a significant challenge in FER in general. 
Models based on 2D FER can be more effectively trained to handle these challenges because there are available data samples that include occlusions. 
3D-based FER databases typically lack such samples, with only the Bosphorus database~\cite{savran2008bosphorus} including data with such occlusions.
In this work, we bridge the gap between 2D and 3D FER by proposing a method to generate high-quality 3D pointclouds from existing 2D FER datasets, enabling \textcolor{blue}{privacy-aware} and robust emotion recognition without relying on costly 3D scanning setups.

\subsection{Privacy Enhancements of 3D FER}
The use of \ac{2D} facial images for \ac{FER} raises significant privacy concerns due to the risk of identity disclosure, unauthorized surveillance, and the misuse of high-resolution biometric data. Global regulations, including the EU AI Act, GDPR, and UNESCO AI ethics recommendations, classify facial images as sensitive personal data, mandating strict compliance for their storage, processing, and transmission~\cite{butt2024analytical,morandin2023ten}.

Recent studies have shown that 2D facial landmarks can be extracted from images and used for FER without storing full facial images~\cite{qiu2019facial,haghpanah2022real}. However, while on-device extraction of 2D facial landmarks can reduce privacy risks compared to full-image FER, these reduced representations still encode biometric characteristics that enable re-identification~\cite{pantic2004facial,haghpanah2022real}, thereby remaining subject to the same privacy vulnerabilities and legal constraints as other biometric identifiers.

As an alternative, 3D pointclouds inherently offer \textcolor{blue}{privacy-aware advantages at the representation level}.\textcolor{blue}{ Their lack of texture and color information reduces the exposure of directly identifiable visual cues and makes them less susceptible to direct identity disclosure.} Unlike 2D images, which contain fine details such as skin tone, hair color and other identifiable features, pointclouds primarily encode spatial geometry. Research confirms that the removal of texture data can complicate identity reconstruction, aligning with modern privacy-by-design principles~\cite{wu2021bioface}.
Despite the advantages, 3D data is not immune to privacy risks. Prior work on 3D Structure-from-Motion (SfM) has shown that even sparse pointclouds can be inverted to recover recognizable scenes~\cite{pittaluga2019revealing}. This indicates that 3D representations can still carry latent identity cues that may be exploited under adversarial conditions.

Several methods can further complicate de-anonymization attempts while maintaining task-relevant information. In the context of \ac{JCAS} embedded in wearables, on-device identity-standardizing affine transformations (such as normalizing inter-ocular distance, nose length, or facial depth) can preserve affective cues while removing individual biometric markers. Prior work has shown that automatic detection of eyes and nose followed by geometric alignment can normalize 3D facial data~\cite{Wu2014AutomatedFaceExtraction} and that pose-normalization via geometry-analysis is feasible for 3D head models~\cite{bevilacqua20093d}. These landmark- and PCA-driven normalization techniques could similarly be adapted to normalize 3D pointclouds to a standard template, thereby reducing identifiable variations while retaining the structural and affective information necessary for emotion recognition. These transformations can be combined with spatial downsampling or partial anonymization techniques, which have been shown to make reconstruction and re-identification substantially more difficult~\cite{wu2021bioface}. Such practices align with privacy-by-design principles and provide resilience against data breaches.

\begin{table*}[!t]
\centering
\caption{Privacy--utility positioning of representative FER paradigms. This table provides a high-level design-space positioning rather than a performance leaderboard; reported utility trends depend on dataset, protocol, and (for DP) privacy budget.}
\label{tab:privacy_utility_positioning}
\renewcommand{\arraystretch}{1.15}
\begin{tabular}{p{1.5cm} p{2.7cm} p{3.0cm} p{2.0cm} p{2.5cm} p{2.8cm}}
\hline
\textbf{Paradigm / approach family} & \textbf{Primary privacy mechanism} & \textbf{Data representation (stored/processed)} & \textbf{Formal privacy guarantee?} & \textbf{Typical deployment burden} & \textbf{Compatibility with our pipeline} \\
\hline
Standard 2D FER (RGB) & None (full texture exposure) & Raw RGB facial images & No & Moderate (standard training/inference) & Not applicable (different privacy objective) \\
Landmark-based FER & Representation minimization (sparse geometry cues) & 2D landmarks / keypoints & No (biometric leakage possible) & Low & Compatible (alternative or auxiliary input) \\
AU-based FER & Representation minimization (semantic/physiological coding) & Action Unit vectors / intensities & No (depends on pipeline) & Low--Moderate (AU extraction + training) & Compatible (can be fused or used as auxiliary) \\
Federated learning FER & Distributed training (data stays local) & Local training + model updates (gradients/weights) & No (unless combined with DP/secure aggregation) & High (protocol, comms, orchestration) & Orthogonal; can be combined with ours \\
Differential privacy (DP) FER & Noise injection during training & Noisy gradients/updates; privacy budget $\varepsilon$ & Yes (under DP assumptions) & High (privacy budget selection, tuning, utility trade-off) & Orthogonal; can be combined with ours \\
Geometry-only 3D FER (this work) & Representation minimization (no texture) & Texture-free 3D pointcloud / mesh geometry & No & Moderate (3D processing + FER model) & Core approach (representation layer) \\
\hline
\end{tabular}
\vspace{-3.5mm}
\end{table*}

\textcolor{blue}{To place this contribution in context, Table~\ref{tab:privacy_utility_positioning} summarizes representative privacy-preserving FER paradigms (e.g., landmark/AU-based inputs, federated learning, differential privacy) and contrasts their privacy mechanism, deployment requirements, and complementarity with our representation-level approach.}
\textcolor{blue}{Approaches such as federated learning and differential privacy target privacy at the training and update levels (distributed optimization and noise injection), whereas our contribution targets privacy at the representation level by reducing the exposure of raw texture-bearing imagery. These paradigms are therefore orthogonal and can be composed: for example, a model trained on geometry-only inputs can additionally be trained with differential privacy or via federated learning if the deployment scenario requires it.}

\textcolor{blue}{\noindent\textbf{Threat model and non-goals:} We consider scenarios where FER must be performed under stringent privacy and regulatory constraints, in which raw RGB facial images are undesirable to store, transmit, or centrally process. Our approach reduces exposure by design by operating on texture-free 3D geometry. However, geometry-only representations may still carry biometric signals and can, in principle, support partial identity inference.}
\textcolor{blue}{By leveraging 3D pointclouds obtained from \ac{JCAS}-enabled sensing, our approach reduces texture exposure by design while maintaining accurate \ac{FER}, supporting regulation-compliant deployments where storing or transmitting raw facial imagery is undesirable.}

\section{Related Works}
\label{sec:related_works}

\subsection{Face Reconstruction from Images}

Reconstructing 3D models from 2D images is a key research direction for overcoming the inherent limitations of flat imagery. 
By inferring 3D geometry, systems gain a more robust and viewpoint-independent understanding of facial structure and expression. 
This capability enables applications in realistic avatar creation, animation and AR/VR applications, while also supporting medical, biometric, and forensic use cases.

The reconstruction of 3D facial structures from 2D images is another pivotal element in FER systems, traditionally achieved through statistical models like 3D Morphable Face Models (3DMM), which use \ac{PCA} to fit face shapes and expressions. Key methods include BFM, which separates shape and expression parameters using PCA~\cite{paysan20093d}; FLAME~\cite{li2017learning}, which enhances reconstruction detail by incorporating real-expression deformation variables~\cite{li2017learning}; and DAD-3dHeads, which further refines this process by regressing FLAME parameters from a wild dataset for improved accuracy in diverse conditions~\cite{martyniuk2022dad}.

Recent state-of-the-art methods are able to achieve 3D facial reconstruction from \textit{in-the-wild} 2D images. 
RingNet~\cite{sanyal2019learning} tackles this problem without requiring paired 2D-to-3D supervision by learning a mapping from 2D images to 3D face representations using identity-labeled images and a consistency loss. 
During training, the network leverages multiple images of the same person to enforce similarity in their latent embeddings while keeping embeddings of different individuals distinct. 
These embeddings are then decoded into a full 3D face model using FLAME~\cite{li2017learning}. 
Building on a similar principle, DECA~\cite{feng2021learning} extends this approach to capture details such as expression-dependent wrinkles. Using a consistency loss, DECA jointly learns a geometric detail model and a regressor that predicts both image-specific and subject-specific parameters. 
The detail model is then used to refine FLAME’s geometry. 
EMOCA~\cite{EMOCA:CVPR:2022} focuses on the geometric changes critical for emotion perception. 
It builds on DECA by adding a trainable branch for facial expression prediction while keeping the rest of the architecture fixed.

Based on~\cite{vilalta2024ai}, we leverage a FLAME-based method~\cite{li2017learning} for 3D face reconstruction from RGB images. 
This choice is motivated by its in-house availability and due to the fact that our goal is not to strictly optimize 3D model generation itself, but to evaluate the end-to-end pipeline, from 2D in-the-wild images to 3D pointcloud reconstruction and subsequent FER inference on \ac{JCAS}-like data.
Also, FLAME’s parametric representation enables integration with models that capture fine-grained details and expression-dependent deformations, as demonstrated by DECA and EMOCA.

The transition from 2D RGB images to 3D pointclouds in the leveraged method requires the use of precise depth estimation. 
Several state-of-the-art estimators have been considered for this purpose. MiDaS~\cite{ranftl2020towards} integrates diverse training datasets and loss functions to optimize depth estimation accuracy, while AdaBins~\cite{bhat2021adabins} introduces adaptive binning for improved performance in complex scenes. 
ZoeDepth~\cite{bhat2023zoedepth} unifies metric and relative depth estimation to deliver practical metrical values. 
Boost Your Own Depth~\cite{miangoleh2021boosting} merges high-resolution estimates to enhance the detail and consistency of depth maps.
Following the findings of~\cite{vilalta2024ai}, this study adopts a combination of ZoeDepth and Boost Your Own Depth for depth estimation. 
Although this approach entails higher memory consumption~\cite{bhat2023zoedepth, miangoleh2021boosting}, it offers optimal accuracy in depth map detail and metrical values~\cite{vilalta2024ai}.

\subsection{3D FER benchmarks}
\ac{DL} models serving as benchmarks for 3D FER have several common characteristics. The main ones are the incorporation of attention~\cite{lin2020orthogonalization, zhu2019discriminative, zhu2022cmanet, sui2022afnetm} as well as feature fusion mechanisms~\cite{lin2020orthogonalization, zhu2022cmanet, sui2022afnetm}. The DA-CNN~\cite{zhu2019discriminative} model introduces an attention module to CNN models. The AFNet-M~\cite{sui2022afnetm} model includes mask attention modules targeting salient facial regions. The FE2DNet and FE3DNet~\cite{lin2020orthogonalization} models employ an orthogonal loss-guided feature fusion approach to ensure that features are orthogonalized before fusion to avoid redundancies. CMANET~\cite{zhu2022cmanet} incorporates a homo-modal curvature-aware attention module that uses a soft mask to guide attention to important facial areas, coupled with a multi-modal (2D+3D) attention fusion module that allows for pixel-level feature interaction \textcolor{blue}{and} feature interaction over a larger field of view. 

The DrFER approach~\cite{li2024drfer}, rather than employing attention mechanisms, uses the concept of disentanglement to address the challenge of intra-class variability in FER. In this context, disentanglement refers to the separation of identity-specific information from affective information. The DrFER model accomplishes this through two main components: a disentangling component and a fusion component. The disentangling component employs a dual-branch architecture to separately learn features associated with facial expressions and identity, thereby creating de-identified and de-expressed versions of the faces. Subsequently, the fusion component recombines these features in a crossover fashion, reconstructing the original face. Throughout this process, various losses are employed to guide the learning of the network effectively. 

The AFNet-M~\cite{sui2022afnetm} and DA-CNN~\cite{zhu2019discriminative} models are examples of how transfer learning is applied using pre-trained 2D architectures to handle 3D data. Both of these models begin with architectures that were originally trained on 2D images, specifically in contexts that are unrelated to FER. The AFNet-M~\cite{sui2022afnetm} model is an adaptive fusion network (2D + 3D) that employs dual-branch ResNet18s~\cite{he2016deep}, which are initialized with pre-trained parameters from ImageNet~\cite{deng2009imagenet}; to enable treating 3D depth images as if they were standard 2D images, they are preprocessed and resized to dimensions of 3$\times$224$\times$224. The DA-CNN model~\cite{zhu2019discriminative} uses five pre-trained CNN models, specifically VGG16-BN~\cite{simonyan2014very} (meaning VGG16 with Batch Normalization (BN) after each convolutional layer), to process the five types of shape attribute maps that represent the 3D facial scan. The VGG-16 models are equipped with bottom-up top-down feedforward attention modules. 

In summary, attention mechanisms, feature fusion mechanisms, and the incorporation of pre-trained 2D-based architectures are key factors contributing to the success of these models in 3D FER tasks. In this work, PointNet++~\cite{qi2017pointnetdeephierarchicalfeature} was selected as the backbone model for 3D FER (cf., Section~\ref{sec:pointnet}). Unlike the aforementioned architectures, PointNet++ does not utilize explicit attention mechanisms, feature fusion, pre-trained 2D networks, or disentanglement. The rationale for its selection is not to achieve state-of-the-art performance, but rather to test whether the 3D data we generated is suitable for FER. PointNet++ is an established benchmark in pointcloud processing, capable of learning hierarchical local and global features directly from raw 3D data. Its relatively simple and modular architecture makes it computationally efficient and practical for deployment on edge devices, such as smart glasses, without the overhead associated with multi-branch attention networks or 2D pre-trained models. As such, it provides a reliable, \textcolor{blue}{replaceable} baseline for evaluating our generated 3D facial scans \textcolor{blue}{within a feasibility-first, end-to-end evaluation scope}, while remaining flexible for future experimentation with alternative architectures.
\textcolor{blue}{Importantly, our contribution is not a new 3D FER backbone; rather, it is the dataset generation and end-to-end pipeline that enables 3D FER learning from in-the-wild 2D data.}

\subsection{THz-based HFWS for 3D Reconstruction}

Usage of high-frequency \ac{EM} waves in the \ac{THz} regime has a potential of simultaneously enabling both \ac{3D} imaging of the human face and the high-throughput communication from the users' devices to the more powerful edge and cloud processing infrastructure.
\ac{3D} imaging using high-frequency wireless waves is an established technology~\cite{yi2023photonic}, which finds its utility in medical imaging~\cite{arnone1999applications}, tomography~\cite{zhang2004three}, and cancer diagnosis~\cite{rahman2016early}. 
Primarily due to the availability of large bandwidth at THz frequencies, \ac{3D} \ac{THz} imaging systems are known to feature sub-millimeter imaging reconstruction accuracy, making them suitable for facial imaging considered in this work~\cite{yi2023photonic,ding2013thz}. 

Promising \ac{THz} \ac{3D} imaging system designs and implementations are based on \acp{SAR}~\cite{ding2013thz}.
The basic concept of radar techniques is to measure the time delay between the transmitted and received \ac{EM} signals, which can then be converted to distance based on the known propagation velocity of the \ac{EM} waves.
In \ac{SAR} systems, a \ac{3D} image can be obtained by synthesizing a \ac{2D} aperture and azimuth \textcolor{blue}{distinguishing} by array signal processing approaches, mostly a digital beamformer~\cite{cumming2005digital}. 
\ac{SAR} systems can be categorized based on the pulse generation method as pulse and \ac{CW} radars. 
Pulse-based radars feature easier implementation at high frequencies and support large bandwidths at the considered small imaging distances of a few centimeters~\cite{abadal2019media}. 
They can be implemented in an energy-efficient manner through graphene-based transceiver implementation~\cite{abadal2015time}. 
As such, we consider pulse-based \ac{SAR} radars as the prime candidate for 3D imaging for \ac{FER} purposes.

Communication using \ac{THz} frequencies is expected as a key component of the 6G communication systems~\cite{jiang2024terahertz,tripathi2021millimeter}.
THz communication at shorter ranges is expected to be enabled through the utilization of pulse-based \ac{MCS} schemes due to the ease of their implementation in small physical form factor suitable for wearable devices such as glasses~\cite{lemic2021survey}. 
The reuse of pulse-based schemes in array-based THz transceiver implementations in the glasses for both 3D imaging of the human face and the high-throughput communication of the sensed data for cloud processing can be considered as an example of a \ac{JCAS} system~\cite{wu2022sensing}.
From the energy consumption perspective, the re-utilization of the same technology \textcolor{blue}{for both} communication and sensing in the \ac{JCAS} systems represents a unique design advantage over the utilization of different technologies for sensing and communication~\cite{wu2022sensing}. 
\textcolor{blue}{In addition to hardware and energy benefits, such HFWS-based 3D imaging supports privacy-aware deployments by avoiding the acquisition and transmission of raw texture-bearing facial imagery.}
As such, we consider the small-scale THz-based \ac{JCAS} system as the prime candidate for meeting the communication and 3D imaging requirements of the considered system.


\section{System Overview}
\label{sec:system_overview}

\subsection{System Design Requirements}

\paragraph{\textcolor{blue}{Privacy-aware} FER}
The system should incorporate \textcolor{blue}{privacy-aware, representation-level mechanisms} by utilizing on-person sensors that employ the \ac{JCAS} paradigm for capturing high-frequency electromagnetic waves to generate detailed 3D facial images. 
This design choice aligns with the global regulatory initiatives for ethical usage of \ac{AI}, with emphasis on protecting fundamental rights, particularly users privacy~\cite{butt2024analytical,morandin2023ten}. 
Taking \ac{EU} legislation as an example, the \ac{EU} \ac{AI} Act outlined in \textit{Title II, Chapter 2, Article 9}~\cite{EUAIAct} mandates that \ac{AI} systems must respect fundamental rights and avoid creating risks to privacy and data protection. By ensuring that the system does not rely on traditional 2D image data, which could pose significant privacy risks due to potential misuse or unauthorized access, it complies with the Act’s requirements to minimize the risk of data misuse (\textit{Title III, Chapter 1, Article 10}~\cite{EUAIAct}). \textcolor{blue}{We use the term \emph{privacy-aware} to denote reduced exposure of directly identifiable visual cues by avoiding texture-bearing imagery; we do not claim formal guarantees against re-identification, inversion, or reconstruction attacks.}
While we consider high-frequency \ac{JCAS} to be the most promising potential enabler for obtaining 3D facial images, it is just one of several potential methods. Other technologies, such as structured light or time-of-flight sensors, may also be considered depending on the specific application requirements and privacy constraints, ensuring flexibility while adhering to the Act’s standards.

\paragraph{Cost-Effective Training}
To comply with the \ac{EU} \ac{AI} Act's requirement for high-quality, reliable, and bias-free training data, as emphasized in \textit{Title III, Chapter 1, Article 10}~\cite{EUAIAct}, the system should avoid relying on expensive and complex training processes based solely on 3D pointcloud data. Instead, the system should be designed to leverage 2D images, which are more readily available from public databases, by transforming these 2D images into corresponding \ac{3D} pointclouds for training purposes. This innovative approach allows the system to utilize widely accessible 2D \ac{FER} datasets, transforming them into 3D data that can effectively train deep learning models like PointNet++. By refining the generated \ac{3D} pointclouds to focus on the facial region, the system can achieve high accuracy while maintaining a cost-effective training process. This approach meets the \ac{EU} \ac{AI} Act’s requirements for using high-quality data and ensuring the system’s economic viability in high-risk \ac{AI} applications like \ac{FER} (\textit{Title III, Chapter 2, Article 13}~\cite{EUAIAct}).

\subsection{System Design}
\begin{figure}[!t]
\centering
\includegraphics[width=\columnwidth]{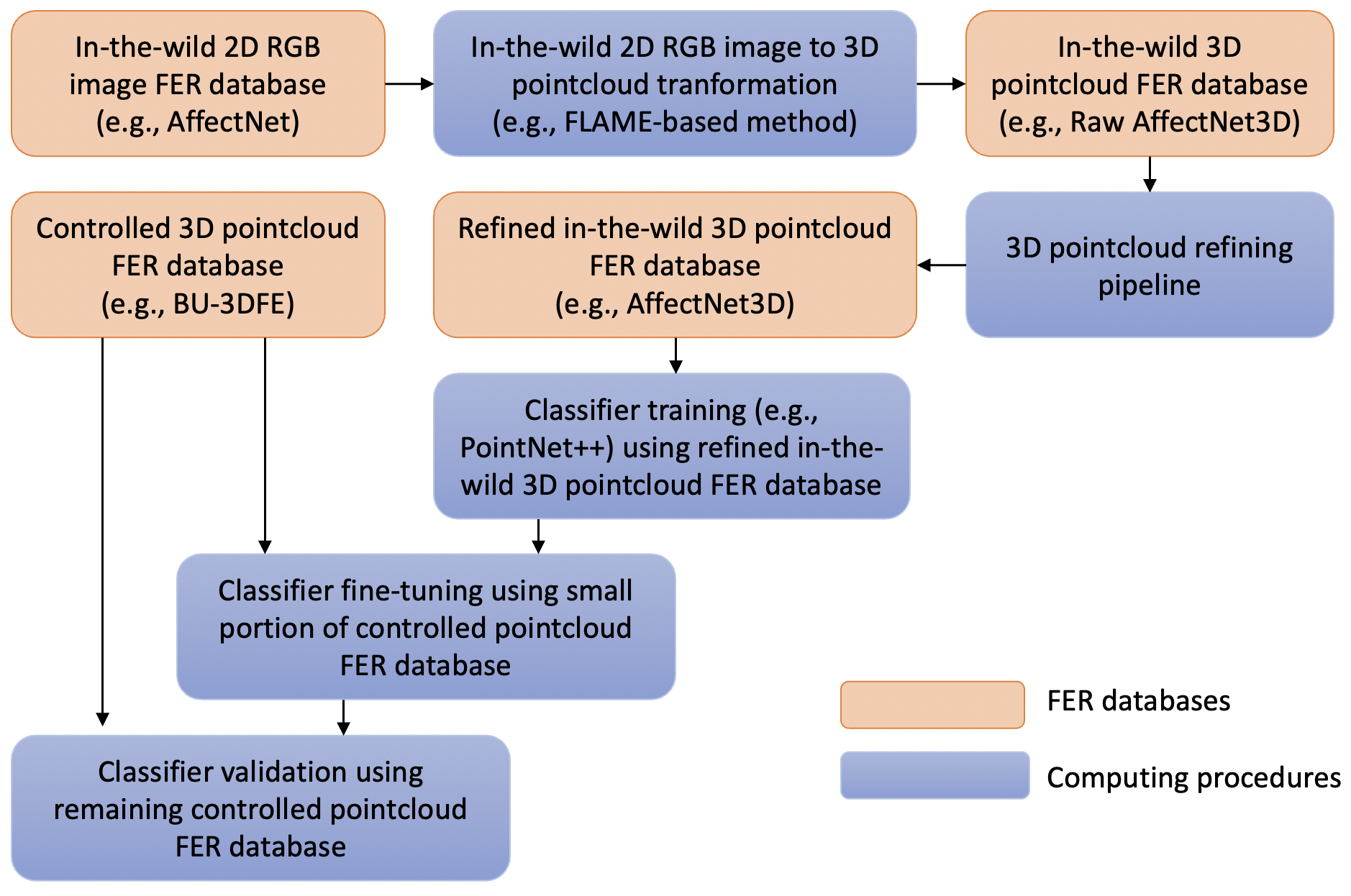}
\caption{System design of the envisioned \textcolor{blue}{privacy-aware} 3D FER system trained on in-the-wild 2D images}
\label{fig:system_design}
\vspace{-4mm}
\end{figure}

This study hypothesizes that using publicly available 2D image data can be leveraged to train a 3D pointcloud \ac{DL} model to obtain similar FER accuracy as \ac{SotA} models that use 2D images as input. 
Figure~\ref{fig:system_design} illustrates the overall system design flow for assessing the validity of the hypothesis.
The target datasets are the \ac{3D} version of the AffectNet database, generated through a data preprocessing method proposed in this work and the BU-3DFE dataset. 
Our methodology follows a three-step approach typical in \ac{FER} studies, starting with custom data preprocessing, and subsequently performing feature extraction and classification. 

First, a method for transforming 2D \ac{RGB} \ac{FER} images into \ac{3D} pointclouds is leveraged to allow the creation of accurate 3D facial representations from 2D data. The generated 3D AffectNet pointclouds \textcolor{blue}{consist} of entire head models with varying expressions. However, popular 3D FER datasets such as BU-3DFE and Bosphorus focus solely on facial structures. This discrepancy led us to question whether the inclusion of non-facial areas could negatively impact the network’s learning.

Second, to improve the network’s capacity to learn from the data, a secondary processing branch was incorporated into the system. This branch refines the pointclouds by isolating the facial region and removing the head and neck, ensuring that only information pertinent to the emotion recognition task is provided to the network. 
Feature extraction and classification are then performed through a single end-to-end training procedure using the selected classifier.

Additionally, we investigate fine-tuning the trained model on a small subset of the BU-3DFE database, a well-established 3D benchmark, to evaluate the effectiveness of leveraging FLAME-generated 3D data for training models in 3D-based emotion recognition.

\subsection{Generation of Raw AffectNet3D}

The \ac{FLAME} method for transforming 2D RGB images into \ac{3D} pointclouds starts by regressing 2D landmarks from the images, which include labeled facial expressions. These 2D landmarks are back-projected to create corresponding \ac{3D} landmarks, aligning them with the \ac{FLAME} template model. The parameters of the \ac{FLAME} model, including identity, expression, and pose, are then optimized to fit the facial data.

To enhance the accuracy of the \ac{3D} pointcloud generation, we utilized ZoeDepth combined with Boost Your Own Depth. This approach integrates high-resolution depth estimation and adaptive binning techniques, improving the detail and precision of the depth maps. The combined method generates high-resolution \ac{3D} pointclouds by sampling the surface of the \ac{3D} facial mesh, which captures fine details of facial geometry and expressions.
The resulting \ac{3D} pointclouds are subsequently refined through grouping, filtering, and subsampling to prepare them for deep learning models. These processed pointclouds are then used for feature extraction and classification in facial emotion recognition tasks.

\begin{figure*}
\centering
\begin{minipage}{.54\textwidth}
  \centering
  \includegraphics[width=\linewidth]{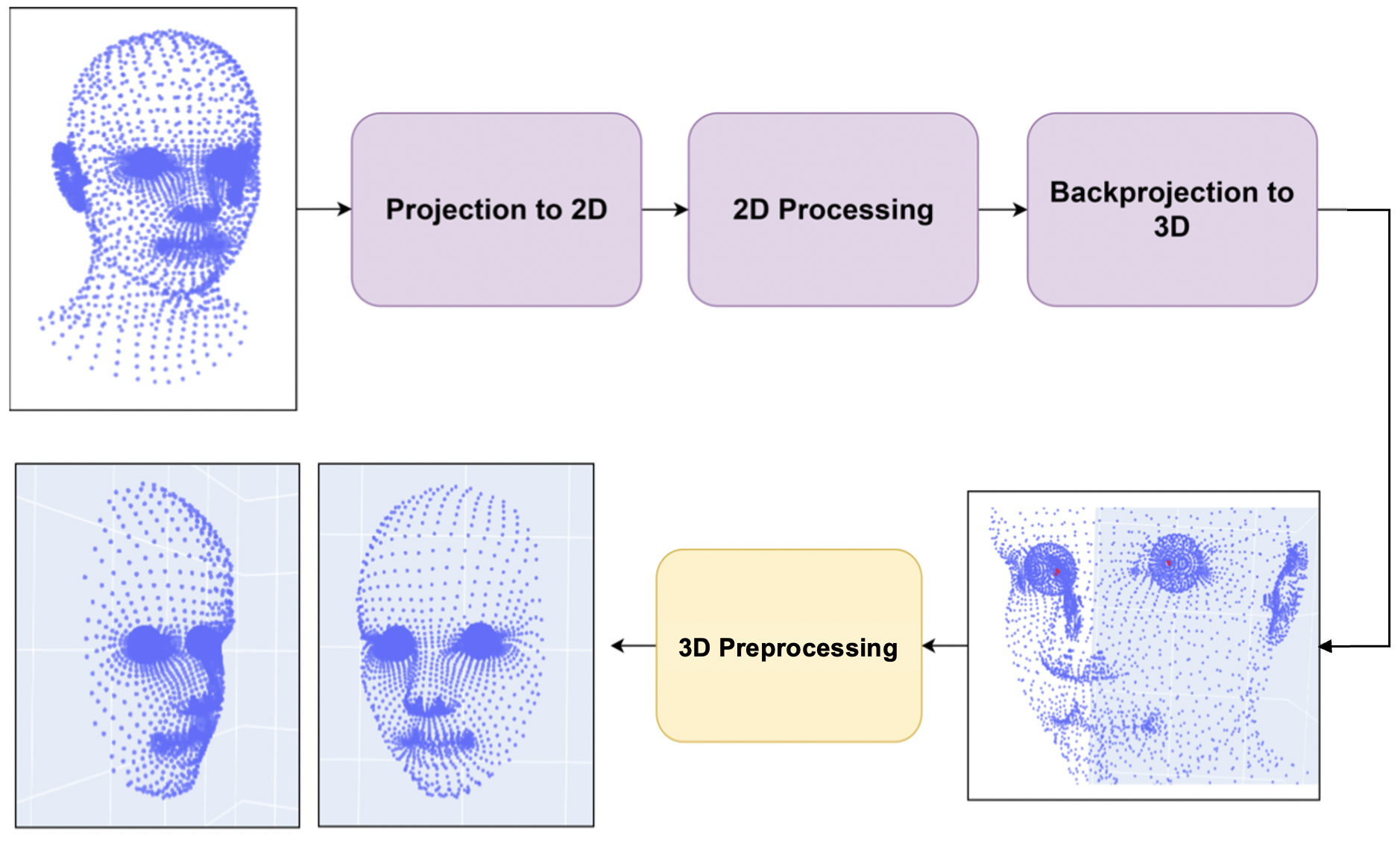}
  \vspace{-2mm}
  \captionof{figure}{Pipeline for refining AffectNet-3D pointcloud, divided into 2D (purple) and 3D (yellow) processing segments}
  \label{fig:refining_pipeline}
  \vspace{-3mm}
\end{minipage}%
\hfil
\begin{minipage}{.41\textwidth}
  \centering
  \includegraphics[width=\linewidth]{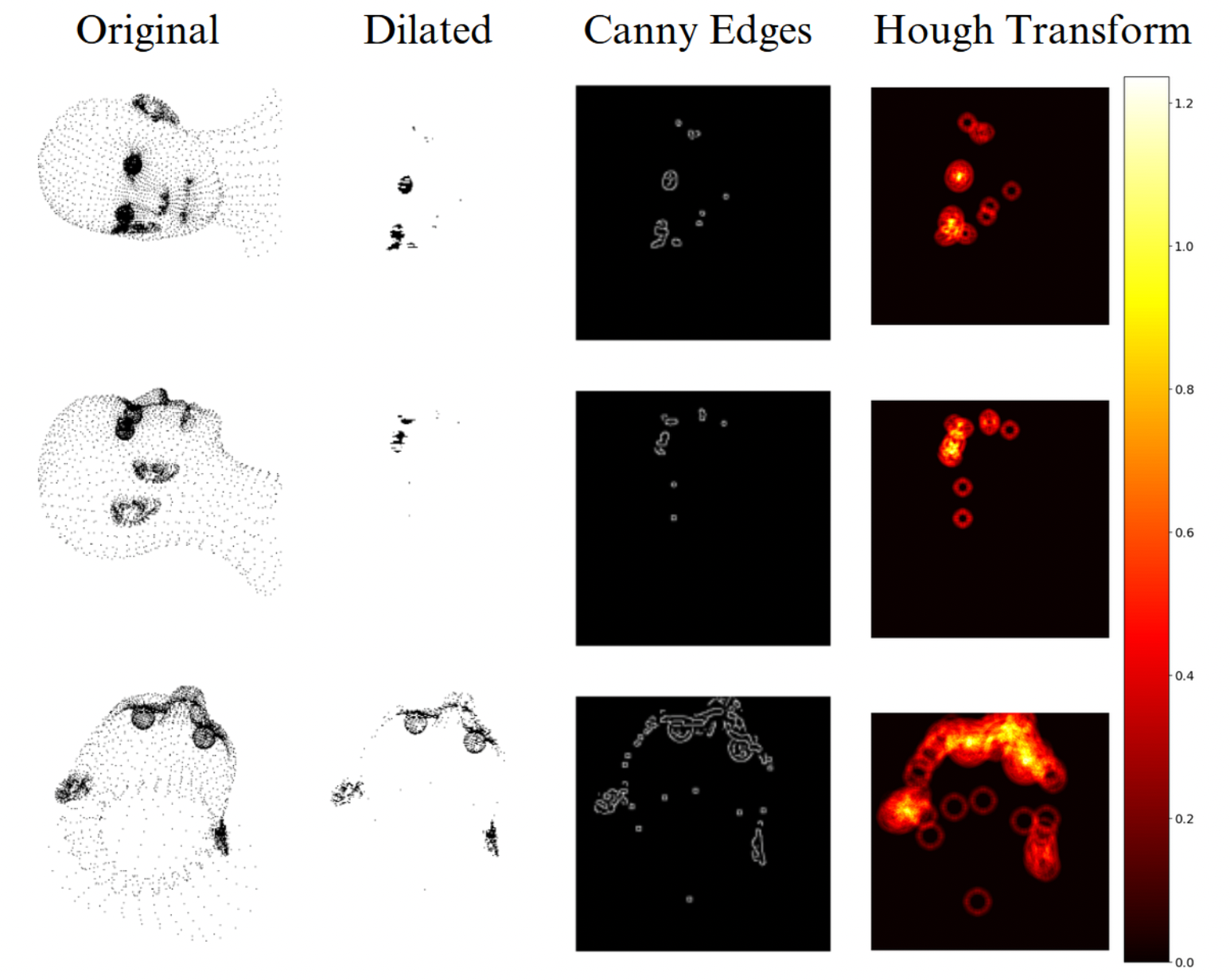}
  \vspace{-2mm}
  \captionof{figure}{Transformations applied to pointcloud projections for eye center localization}
  \label{fig:pipeline_example}
  \vspace{-4mm}
\end{minipage}
\end{figure*}

\subsection{Raw AffectNet3D Refining}
\label{sec:database_refining}

Due to the fact that the generated pointclouds contain entire head models, the model receives substantial amounts of non-essential information as input. 
Only the facial regions of a pointcloud contain expressions, the remaining portions provide no useful data for the FER task. 
This work hypothesizes that as a result this could ultimately hinder the model's ability to recognize facial expressions. Moreover, the presence of a significant amount of irrelevant points could lead to unnecessary computational overhead and prolonged processing times.  
With the aim of improving the network’s learning ability from the data, a data refining module is envisioned in our system design, as illustrated in Figure~\ref{fig:system_design}. 

The module is envisaged to refine the pointclouds by isolating the facial region, excluding the head and neck.
This is done using the model’s eye orbits as a reliable anatomical landmark for establishing a cropping boundary.
All AffectNet3D pointclouds are built with the (0,0,0) coordinate point consistently positioned at the center of the head.
Nevertheless, the orientation of the heads varies significantly: they display different degrees of rotation and tilt.
The method we developed takes into account the angular orientation of the head when defining the cropping plane to isolate the facial region. 
Our pipeline is designed to detect the center of the eye orbits, establish a cropping plane at a calculated distance of 2.3 times the orbit radius, and execute a vertical crop of the pointcloud. 
Empirical evaluations of this technique have proven its efficacy in consistently isolating the facial region.

The overall data refining pipeline proposed in this work is depicted in Figure~\ref{fig:refining_pipeline}. 
This pipeline is divided in two major parts: i) A 2D processing segment aimed at identifying the centers of the eyes and determining the radius of the eyeballs, and ii) a 3D processing segment focused on defining the cropping plane and filtering the pointcloud accordingly. 

\subsubsection{2D Processing Segment} The initial step in this segment involves projecting the pointcloud onto a 2D plane using the three coordinate axes.
Subsequently, the projections undergo a series of transformations that result in the identification of the three-dimensional coordinates of the eye centers, as well as their radii. 
The transformations consist of:

\begin{enumerate}
 \item Converting the 2D projections into binary image format enables the application of image processing methods. Based on empirical evidence, only the x-y and z-x projections are used, as they have proven to be more effective in simplifying the identification of eye positions. 
 \item Applying mathematical morphology~\cite{serra1983image} for easier eye region identification. Specifically, the dilation transformation is used to ’open’ the structures within the image. In a binary image where dense, black regions represent potential areas of interest (eye orbits in this case), dilation helps to eliminate smaller, irrelevant black regions. The structuring elements (kernels) used are empirically set to 1x2 for the x-z projection and 2x5 for the x-y projection. 
 \item Using Canny Edge Detection~\cite{canny1986computational} to reduce noise by identifying the edges of features in the image, clarifying regions before applying more complex transformations.
 \item Applying the Hough Circle Transform to detect circles within the image that correspond to potential eye orbits. The range of possible radii for the circles is empirically determined. The Hough Circle Transform returns a list of circle centers and their radii.
 \item Filtering the results from the Hough Transform~\cite{hassanein2015survey} by retaining only the circles that most likely represent the eyes. This filtering process considers factors such as the Hough transform confidence and the geometric plausibility of intercircle distances, i.e., the expected distance between two eyes.
 \item Matching the coordinates of eye centers between projections. Once the eye centers are identified in both the x-y and x-z projections, their coordinates are matched. This involves aligning the z-coordinates from the x-z projection with the x and y coordinates from the x-y projection. The correct pairs are identified by ensuring the closest match in the x-coordinates between the two projections. The final step of the 2D processing involves back-projecting the eye  coordinates from 2D to 3D.
\end{enumerate}

\subsubsection{3D Processing Segment} The objective of the 3D processing segment is to isolate the facial area from the rest of the pointcloud. This is achieved by constructing a plane (based on reference points near the eye centers) and using this plane to determine which points to keep or discard. The steps involved are as follows (cf., Figure~\ref{fig:pipeline_example}): 1. Generating Lines from Eye Centers: Lines are drawn from each eye center to the origin of coordinates (0,0,0) which, as already stated, is consistently located at the center of the head. 2) Defining Reference Points for the Plane: Two points, prigth and pleft, are chosen along these lines, positioned at an empirically determined distance of 2.3 times the radius of each eye as detected by the Hough Circle Transform. 3) Plane Construction: The plane is defined by creating two vectors: one connecting pright to pleft and another aligned with the positive Y-axis (which points towards the neck). This configuration ensures that the plane’s normal vector points towards the facial region of the pointcloud. 4) Filtering Points Relative to the Plane: The pointcloud points are filtered based on their position relative to the plane using the inequality
$Ax + By + Cz + D > 0$, where (x, y, z) are the coordinates of a point in the pointcloud, while (A,B,C) are the components of the plane’s normal vector, with D being the perpendicular distance of the plane from the origin.
Points that satisfy this condition lie on the ’positive’ side of the plane, which has been oriented to coincide with the facial region of the pointcloud.
Before feeding the refined pointclouds into the network, they undergo \ac{FPS} to standardize the number of points across all samples. The number of points is adjusted to match the smallest pointcloud in both the test and training sets, which contained 1,496 points. 
In this work, we have considered reducing the pointclouds to 512 centroids, with each centroid having 16 neighbors, due to the fact this number of centroids empirically yielded optimal performance in terms of the trade-off between classification accuracy and training time.
Regardless, the optimization for the number of centroids is considered as out of scope. 

\subsection{Utilized PointNet++ Model}
\label{sec:pointnet}

To process the generated 3D pointclouds, PointNet++~\cite{qi2017pointnetdeephierarchicalfeature} was selected due to its proven performance in the \ac{SotA} literature and the availability of mature open-source implementations. 
Specifically, the PointNet++ is inspired from~\cite{enhancedFERPointnet2021}, in which reasonable accuracies were achieved on two different facial pointcloud databases (Bosphorus~\cite{savran2008bosphorus} 69.01\% and SIAT-3DFE~\cite{siat-3dfe} 78.80\%) using the PointNet++ network architecture. PointNet++ constitutes a foundational design component of our hypothesis evaluation pipeline. 
The primary goal is to establish a baseline for point-based \ac{DL} on 3D FER. This decision does not limit the framework to this specific architecture, and more advanced \ac{DL} models such as voxel-based approaches can be integrated in future work. 

\begin{figure*}[!t]
\centering
\includegraphics[width=0.74\textwidth]{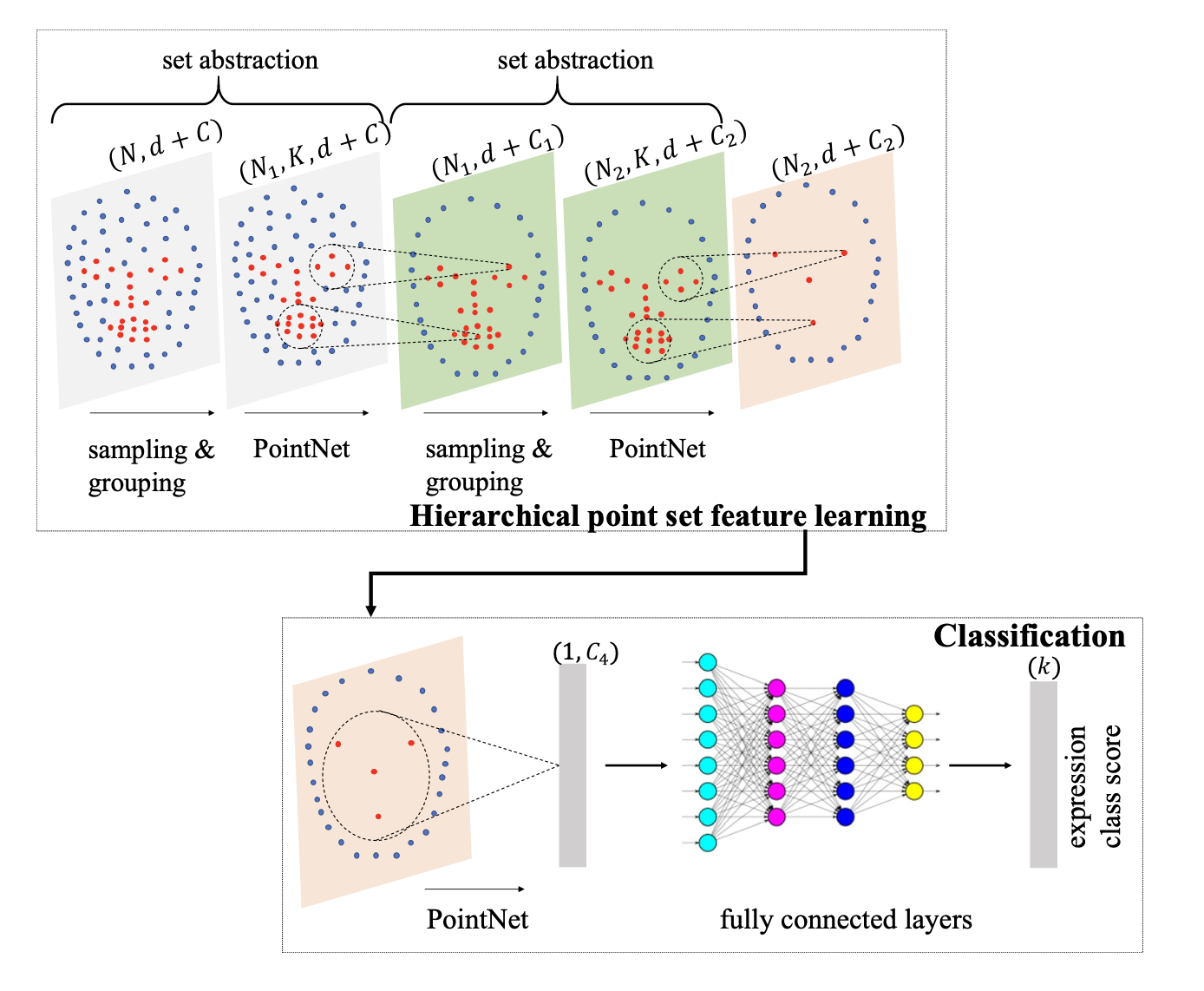}
\vspace{-2mm}
\caption{High-level design of the PointNet++ network, extracted from~\cite{enhancedFERPointnet2021}}
\label{fig:PointNet++}
\vspace{-4mm}
\end{figure*}

The PointNet++ network architecture is given in Figure~\ref{fig:PointNet++}. 
PointNet++~\cite{qi2017pointnetdeephierarchicalfeature} is a geometric deep learning model that can be seen as an extension of the previous PointNet~\cite{qi2017pointnet} architecture with incorporated hierarchical structure. The original PointNet architecture is capable of doing feature extraction using a function that maps a set of points to a feature vector invariant to the input order. In this function, multi-layer perceptron networks are used to learn features independently for each point and global features are extracted using a max-pooling layer. However, the independent learning per point causes the PointNet not being able to capture local context between points in local regions. PointNet++ adresses this problem by adding sampling and grouping layers. A PointNet++ based model, designed for classification of pointclouds, consists of a number of set abstraction levels for feature extraction which are then followed by a classifier consisting of fully connected layers. A single set abstraction level contains three key layers: a sampling layer, a grouping layer and a PointNet layer. The sampling layer samples a number of points from a set of input points, which determines the centroids for local regions in the pointcloud. The grouping layer determines local region sets by finding neighbouring points for each centroid. Then, the PointNet layer encodes these local regions into feature vectors. 

As visible in Figure~\ref{fig:PointNet++}, the PointNet++ model is composed of two set abstraction levels which form the feature extraction component of the model. This is followed by the classifier which consists of a single PointNet layer that converts the sets of points into a single feature vector such that it can be given as input to the following fully connected layers. 


\section{Performance Evaluation Methodology}
\label{sec:methodology}

\begin{figure*}
\centering
\begin{minipage}{.63\textwidth}
  \centering
  \includegraphics[width=\linewidth]{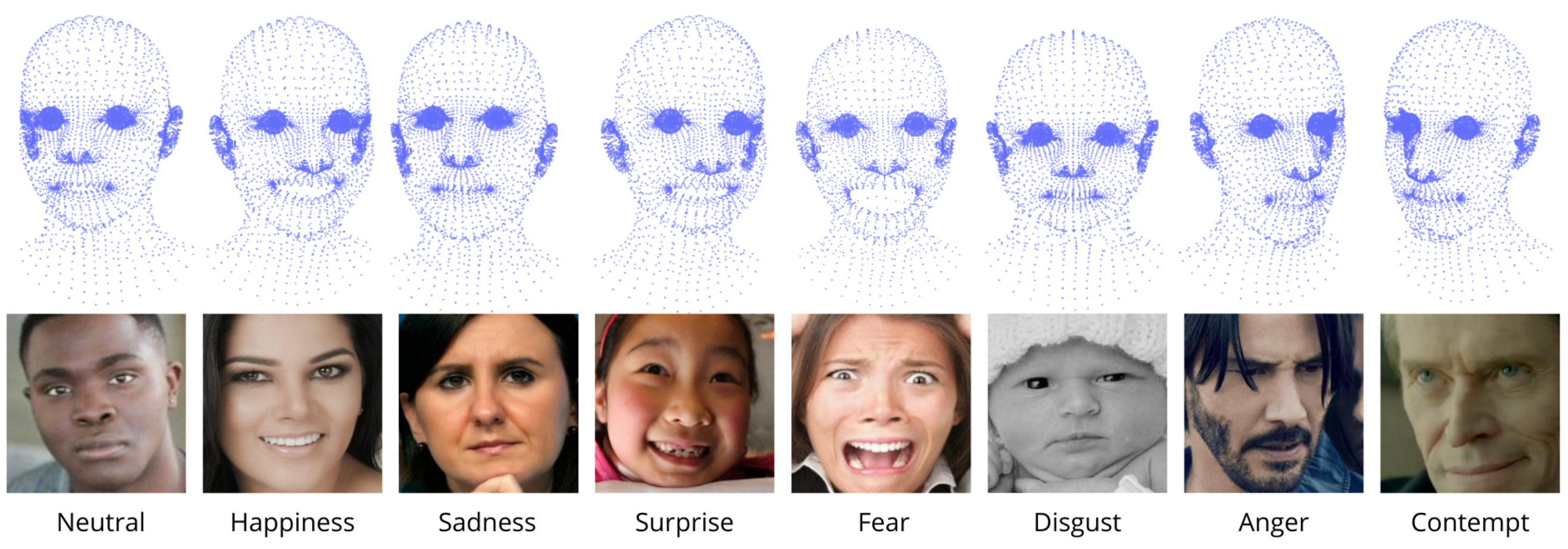}
  \vspace{-1mm}
  \captionof{figure}{Examples of original 2D AffectNet images alongside their corresponding 3D generated pointclouds}
  \vspace{-3mm}
  \label{fig:affectnet3D_example}
\end{minipage}%
\hfil
\begin{minipage}{.32\textwidth}
  \centering
  \includegraphics[width=\linewidth]{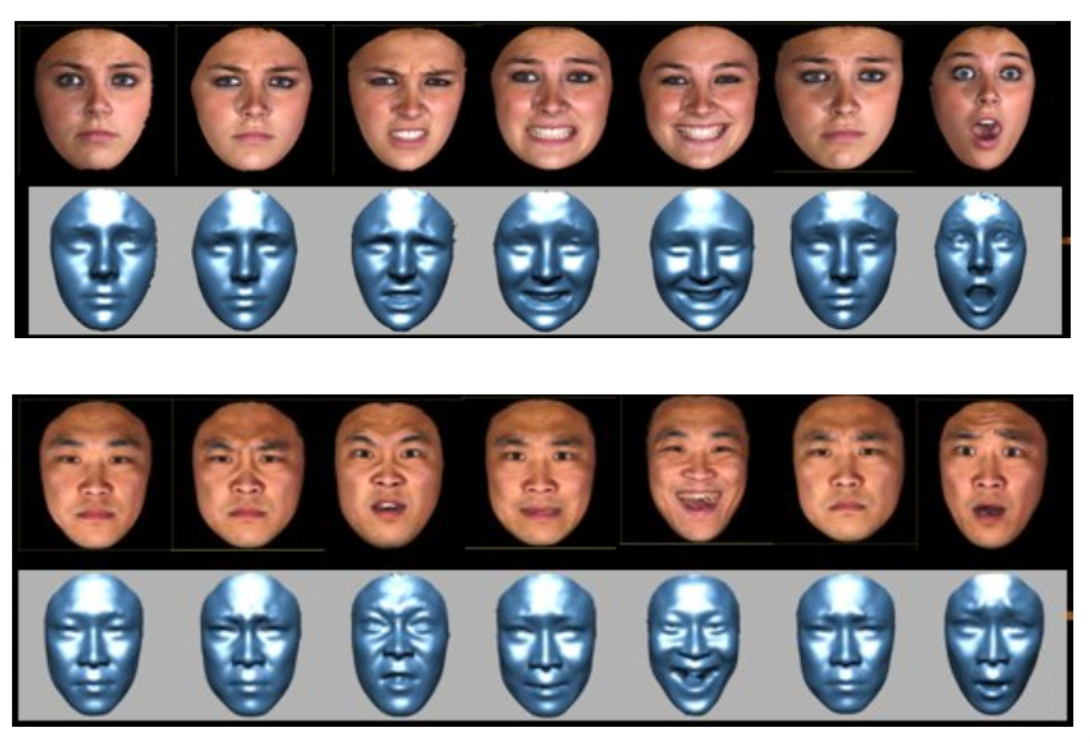}
  \vspace{-1mm}
  \captionof{figure}{Examples of BU-3DFE 3D pointclouds}
  \label{fig:bu-3dfe_example}
  \vspace{-4mm}
\end{minipage}
\end{figure*}

The performance evaluation of the proposed approach is designed to assess the effectiveness of the entire processing pipeline rather than analyzing individual components in isolation.
The absence of corresponding ground-truth pointclouds makes validating the \ac{3D} reconstruction of AffectNet images (cf., Figure~\ref{fig:affectnet3D_example}) inherently challenging. 
Without reference data, it is impossible to directly quantify reconstruction accuracy or compute conventional error metrics, such as point-to-point distances. 
This limitation underscores the importance of evaluating the reconstructed \ac{3D} data within the context of a practical downstream FER task, where performance can be assessed through classification metrics rather than reconstruction errors. 
Accordingly, we qualify the data by its ability to support FER using a standard DL model (PointNet++). 
This model serves purely as a validation tool and could be replaced by a better performing alternative. 
The focus is on demonstrating that the FLAME-based reconstruction method, adapted here for \ac{3D} \ac{FER} to ensure pipeline compatibility while preserving high-fidelity facial geometry, enables accurate and \textcolor{blue}{privacy-aware} \ac{3D} \ac{FER}.
\textcolor{blue}{In this work, \emph{privacy-aware} refers to reducing exposure of directly identifiable cues by relying on texture-free geometry rather than raw facial imagery; we do not claim formal privacy guarantees against re-identification, inversion, or reconstruction attacks.}
Consequently, an ablation study aimed at optimizing individual components is beyond this work's scope.

\vspace{-1mm}
\subsection{Evaluation Setup}

The experimental evaluation incorporates the BU-3DFE dataset (cf., Figure~\ref{fig:bu-3dfe_example}), which provides high-quality \ac{3D} facial pointclouds without the noise typically present in practical \ac{JCAS}-based imaging. The BU-3DFE database has been chosen due to it being one of the most frequent databases for validation in \ac{3D} \ac{FER} studies \cite{alexandre2020systematic}. The purpose of this evaluation is not to demonstrate dataset generalizability, but to validate the reliability of the proposed FLAME-based reconstruction method. Specifically, BU-3DFE serves as a high-quality reference to determine whether the reconstructed \ac{3D} pointclouds from AffectNet (cf., Figure~\ref{fig:affectnet3D_example}) possess sufficient geometric and expressive fidelity to enable effective emotion classification. In this sense, the comparison is used to assess data quality and task relevance rather than cross-dataset generalization.

Furthermore, to assess whether using FLAME-generated \ac{3D} data for \ac{DL} training offers advantages for \ac{JCAS}-based \ac{FER} in continuous monitoring scenarios, beyond relying solely on the readily available \ac{3D} \ac{FER} datasets, we evaluate the trained network on pointclouds where specific facial regions have been masked to simulate realistic wearable sensing conditions.
\textcolor{blue}{We emphasize that this masking analysis evaluates utility under reduced observability and does not constitute a formal privacy-robustness evaluation (e.g., against re-identification or inversion attacks).}

The training datasets used in the following experiments are a modified version of the original 3D AffectNet and the BU-3DFE dataset. 
The training subset of the 3D AffectNet database was reduced from 287,651 samples to 30,000 samples, selecting 3,750 samples for each of the eight emotion categories. 
The filtering process consisted of dataset shuffling to mitigate order bias, that could hinder the model's generalization capabilities, along with sample selection. Regarding BU-3DFE, the dataset was used without any filtering for all experiments. Our experiments present results for both the preprocessed and unprocessed versions of the 3D AffectNet database. In the following sections, we will refer to the preprocessed dataset as AffectNet3D and the unprocessed dataset as Raw AffectNet3D.

Model training sessions were executed using an NVIDIA A100 80GB PCie \ac{GPU} with \ac{CUDA} v.~12.0, within a Miniconda environment running Python 3.7. In addition to this, the system operated using 4 \ac{CPU} with 16GB of \ac{RAM}.

\subsection{Evaluation Scenarios}

Our evaluation scenarios include two sets of experiments. First, we do a set of baseline experiments (E1 - E3). These experiments are meant to be compared to our proposed method and to also serve as sanity checks. Then, we do a set of experiments that evaluate our proposed strategy (E4 - E6). 

\subsubsection{Experiment 1 (E1.a/b)} First, we aim to establish an ``oracle'' solution by training the PointNet++ model on the complete training portion of the BU-3DFE database (i.e., 80\%). 
The model is first trained using the full size of the pointclouds (E1.a). In addition to this, we also train the model on pointclouds that contain 512 points (E1.b) which were obtained through \ac{FPS}.
512 points have been selected empirically for optimally balancing the trade-off between classification accuracy and training time.  
As BU-3DFE is a commonly used FER database, finding this baseline gives us an understanding on the performance of the PointNet++ regarding capturing meaningful pointcloud semantics for FER tasks with regard to the existing literature. 
Both training procedures were done for 100 epochs. 

\subsubsection{Experiment 2 (E2.a/b)} This experiment includes training the PointNet++ model using only 25\% percent of the BU-3DFE database and performing inference using the remaining unseen 75\% percent. Again, we repeat two times, once using the full size of the pointclouds (E2.a) and once using pointclouds downsampled to 512 points (E2.b). This experiment serves as a baseline for comparison with the fine-tuning experiments of the proposed solution (i.e., E6.a and E6.b). 
The training was carried out for 15 epochs, same as the fine-tuning in E6.a/b.

\subsubsection{Experiment 3 (E3.a/b)} The PointNet++ model is trained on an even smaller portion of BU-3DFE.
Specifically, we use only 10\% and perform inference using the remaining unseen 90\% of the database, repeated for full size pointclouds (E3.a) and for pointclouds having 512 points (E3.b). This experiment serves as a baseline for experiment E6.c. Training for both \textcolor{blue}{solutions} was done for 15 epochs.
These experiments were defined to showcase the effects of reducing the amount of fine-tuning data on the classification accuracy. 

\subsubsection{Experiment 4 (E4.a/b)} This experiment includes training the PointNet++ model on the raw AffectNet3D dataset. As before, E4.a stands for using the full sized pointclouds and E4.b for using pointclouds downsampled to 512 points. 

\subsubsection{Experiment 5 (E5.a/b)} Similar to Experiment 2, the PointNet++ model is trained on AffectNet3D. However, in this experiment, AffectNet3D has been preprocessed before training using the procedure described in Section~\ref{sec:database_refining}.

\subsubsection{Experiment 6 (E6.a/b/c)} This experiment is a successive experiment to experiments E4.b and E5.b using the resulting models after training on the downsampled AffectNet3D and downsampled raw AffectNet3D. The trained PointNet++ models are fine-tuned for 15 epochs using only 25\% of the BU-3DFE dataset (E6.a/b). Then, the fine-tuned models are tested using the other-never seen-part (75\%) of the BU-3DFE dataset. In addition to this, this is again repeated for the model trained on the preprocessed downsampled AffectNet3D but now only using 10\% of the BU-3DFE database where the other 90\% is used for inference (E6.c).

\section{Evaluation Results}
\label{sec:results}

Table~\ref{tab:results} displays the classification accuracy and training times observed in all experiments. First, we analyse the results of the baseline experiments (i.e., E1-E3) after which we examine the results of our proposed strategies (i.e., E4-E6) with reference to the baseline results. 
Experiment E1 shows that PointNet++ is capable of learning features present in the facial pointclouds, resulting in meaningful emotion classification. In addition to this, it shows that having less points per pointcloud logically results in a lower training time, but also increases the inference accuracy. Regarding experiments E2 and E3, we see that if PointNet++ is trained on a small portion of BU-3DFE (25\% and 10\%, respectively), it can still achieve inference accuracy considerably better than random.

\begin{table*}[!t]
  \caption{Summarized classification performance observed in different experiments}
  \footnotesize
  \label{tab:results}
    \centering
    \begin{tabular}{l c c c c c} 
    \hline
    \textbf{Experiment/Metric} & \textbf{\makecell{Used part \\of BU-3DFE [\%]}} & \textbf{\makecell{Training \\accuracy~[\%]}} & \textbf{\makecell{Validation \\accuracy~[\%]}} & \textbf{\makecell{Inference \\accuracy~[\%]}}  & \textbf{\makecell{Training \\time~[min]}} \\ \hline
    \multicolumn{5}{c}{\textbf{BASELINE: Training and Inference on BU-3DFE}} \\ 
    \textbf{E1.a:} Raw BU-3DFE (full) & 80 & 96.50 & - & 74.02 & 198 \\
    \textbf{E1.b:} Raw BU-3DFE (512) & 80 & 98.25 & - & 78.90 & 162 \\
    \textbf{E2.a:} Raw BU-3DFE (full) & 25 & 67.34 & 50.12 & 45.57 & 10 \\
    \textbf{E2.b:} Raw BU-3DFE (512) & 25 & 66.13 & 62.98 & 49.12 & 8 \\ 
    \textbf{E3.a:} Raw BU-3DFE (full) & 10 & 84.38 & 50.00 & 37.34 & 4.5 \\
    \textbf{E3.b:} Raw BU-3DFE (512) & 10 & 72.92 & 45.31 & 42.83 & 4 \\ 
    \hline
    \multicolumn{5}{c}{\textbf{PROPOSED: Training on AffectNet3D, Fine-tuning and Inference on BU-3DFE}} &  \\
    \textbf{E4.a:} Raw AffectNet3D (full) & 0 & 42.99 & 43.37 & 15.57 & 2040 \\
    \textbf{E4.b:} Raw AffectNet3D (512) & 0 & 58.40 & 41.54 & 16.76 & 1764 \\
    \textbf{E5.a:} AffectNet3D (full) & 0 & 51.52 & 42.41 & 16.12 & 447 \\
    \textbf{E5.b:} AffectNet3D (512) & 0 & 48.84 & 42.07 & 14.21 & 396 \\ 
    \textbf{E6.a:} Raw AffectNet3D + Fine-tuning & 25 & 78.42 & 69.41 & 69.76 & 2040 + 8 \\
    \textbf{E6.b:} AffectNet3D + Fine-tuning & 25 & 89.31 & 70.25 & 70.64 & 1764 + 8 \\
    \textbf{E6.c:} AffectNet3D + Fine-tuning & 10 & 88.54 & 67.19 & 58.52 & 1764 + 3.5 \\ \hline    \end{tabular}
    \vspace{-3.5mm}
\end{table*}

Experiment E4 demonstrates that direct training on the raw AffectNet3D is possible as it results in training/validation accuracy better than random. However, the features extracted by such training do not transfer to another FER database as is shown by the inference accuracy on BU-3DFE. This is the case for both E4.a and E4.b, yielding low inference accuracies on BU-3DFE with no fine-tuning of 15.57\% and 16.76\%, respectively.
This low inference accuracy is due to the cross-dataset transfer from FLAME-generated, full-head AffectNet3D pointclouds to the studio-scanned, face-only BU-3DFE with no fine-tuning, the pretrained PointNet++ yields low direct inference (16.76\%), whereas a brief fine-tuning on 25\% of BU-3DFE raises accuracy to 70.64\% (E6.b), indicating the features become effective once the domain gap is bridged.

Regarding experiment E5, we train on refined AffectNet3D with the aim to improve accuracy and/or training time. We see that this does not increase the inference accuracy, yet it substantially lowers the training time by four times. 
For all sub-cases of experiment E6, we observe that fine-tuning results in a competitive inference accuracy on BU-3DFE. When comparing to the baseline, the accuracies from E6.a, E6.b and E6.c are higher than the ones of the corresponding baselines where only a small portion of BU-3DFE is used for training (i.e., E2.b and E3.b). 

\begin{figure}
\vspace{-1mm}
\centering
  \includegraphics[width=0.78\linewidth]{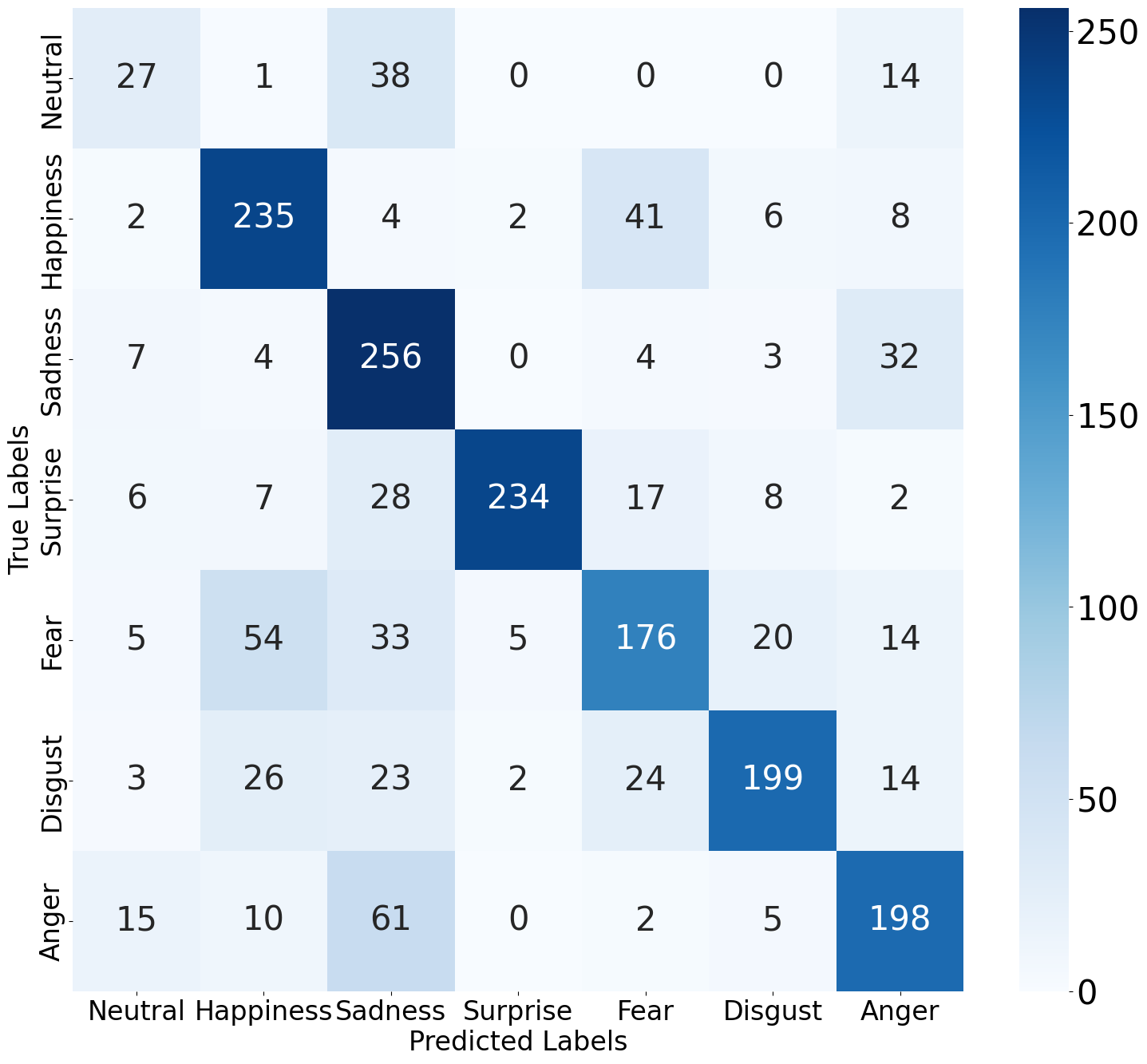}
  \caption{Confusion matrix for experiment E6.b}
  \label{fig:confusion_matrix}
  \vspace{-5mm}
\end{figure}

When considering both inference accuracy and training time, experiment E6.b yields the best results (70.64\% on BU-3DFE). We note that it is only 8.26\% lower than the ones yielded by the oracle solution in experiment E1.b. Figure~\ref{fig:confusion_matrix} displays the confusion matrix for the results of experiment E6.b. 
The model shows notable confusion between "Happiness" and "Fear" with 41 instances of "Happiness" being misclassified as "Fear." Additionally, there is significant confusion between "Sadness" and "Anger"." Further tuning of features or improving the dataset balance may help reduce these pronounced errors.

The main trade-off of the proposed solution is the increased training time compared to the baseline, primarily due to the fact that AffectNet3D is a much larger database than BU-3DFE. 
For example, training on AffectNet3D in E6.a takes 2040 minutes plus 8 minutes for fine-tuning, whereas training on BU-3DFE in E1.a takes only 198 minutes. 
However, given that the training only needs to be performed once, we do not consider this increase in time as a significant drawback. One could envision a generic training of the proposed \ac{FER} system on AffectNet3D, followed by collecting a small sample of user-specific 3D FER pointclouds for fine-tuning to individual users. 
In this scenario, user data collection requirements would be substantially reduced, and the system could be made operational within the fine-tuning latency of just a few minutes.

While the current framework is designed for full-facial pointclouds (e.g., full-head helmets), devices with more limited coverage, such as smart glasses, are more practical for continuous emotion monitoring. Although \ac{3D} pointclouds inherently capture less identifiable information (e.g., color, texture, or lighting) than traditional \ac{2D} images, prior studies show that geometry-only data may still allow partial identity reconstruction~\cite{Chelani2021HowPA, Wang20213DFacePointCloud, YangX2024Coordinate-wise}. Applying geometric transformations or restricting visible facial regions can substantially reduce re-identification risks~\cite{Chelani2021HowPA, YangX2024Coordinate-wise, YangYDigitalmask2022}, suggesting that wearables with limited facial coverage could further enhance \textcolor{blue}{privacy awareness} \textcolor{blue}{by reducing the amount of potentially identifying geometry captured}.
 
To evaluate the system’s adaptability to different wearable configurations, we simulate varying degrees of facial coverage representative of \ac{JCAS}-enabled devices by applying region-based masks to the reconstructed pointclouds, constraining the visible areas to approximate wearable sensing conditions.
\textcolor{blue}{This analysis focuses on utility under reduced coverage and should not be interpreted as an empirical validation of privacy robustness against adversarial attacks.}

\begin{figure}
\vspace{-2mm}
\centering
  \includegraphics[width=0.97\linewidth]{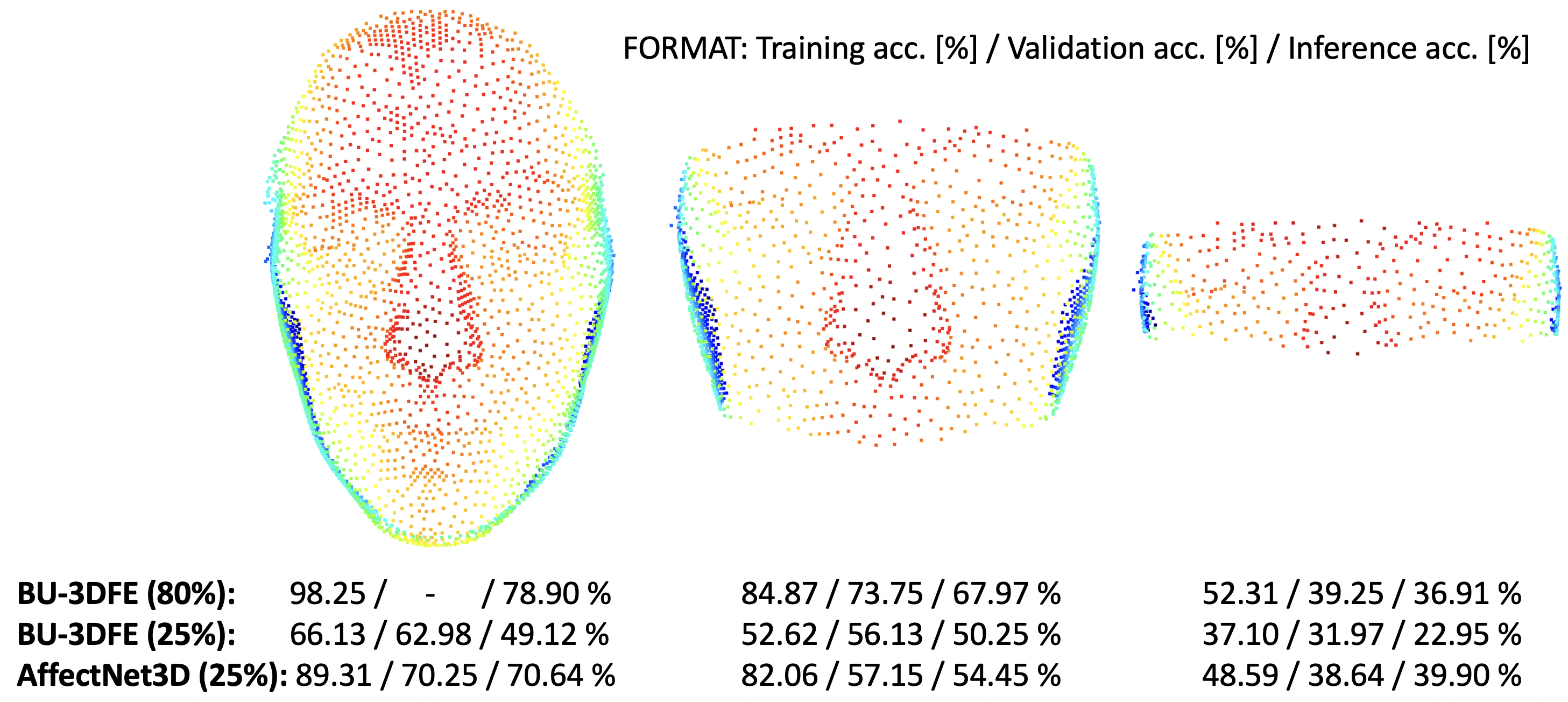}
  \caption{\textcolor{blue}{Classification accuracy under reduced facial coverage for example masking strategies (utility under constrained observability)}}
  \label{fig:masking}
  \vspace{-4mm}
\end{figure}

As shown in Figure~\ref{fig:masking}, two masking strategies were evaluated, reflecting the expected coverage of devices such as smart glasses and \acp{HMD} for \ac{xR} applications~\cite{struye2024toward}. The results indicate that classification accuracy can be to a certain extent maintained under reduced coverage conditions. Specifically, our proposal that combines AffectNet3D with 25\% fine-tuning on BU-3DFE maintains the highest ratio of the classification accuracy compared to the considered oracle (i.e., using 80\% of BU-3DFE for training) and baseline (using 25\% of BU-3DFE for training). These findings establish a promising insight toward expanding the system's applicability to wearables with more constrained facial coverage.


\section{Discussion and Future Work}
\label{sec:discussion}

\subsection{Accuracy Enhancements}
To enhance accuracy, multiple consecutively captured pointclouds could be merged to represent a single emotion. Emotions typically last around 90 seconds, which provides ample time to collect several pointclouds and aggregate them to improve classification accuracy. This compounding approach can potentially refine emotion detection by leveraging temporal data. Additionally, \ac{JCAS} used for 3D structural imaging of the face can be adapted to estimate the temperature of different facial regions, which is a feature that has been demonstrated to aid \ac{FER}~\cite{berlovskaya2020diagnosing}. Integrating thermal information with 3D facial data could enhance the robustness of emotion classification, offering a more nuanced understanding of facial expressions.

While this work focuses on evaluating the use of AffectNet3D as a pre-training dataset with BU-3DFE, incorporating additional databases would help assess and improve the generalizability of the proposed pipeline. Evaluating the inference performance on other datasets could be part of the process towards enhancing the direct transfer accuracy of the pre-trained model without fine-tuning.

Moreover, PointNet++ was used as a baseline model for \ac{3D} \ac{FER} due to its well-established performance in pointcloud-based classification tasks. Future research could explore more advanced \ac{DL} models. 
For instance, \acp{GCN} such as \ac{DGCNN} could better capture spatial relationships between facial landmarks, while voxel-based models like MinkowskiNet could leverage sparse convolutions to enhance feature extraction. 
Additionally, transformer-based architectures, such as \acp{PT}, may provide improved global feature learning and robustness to occlusions. 
Exploring these models could lead to further improvements in classification accuracy, especially when dealing with incomplete or noisy pointclouds expected in real-world \ac{THz} imaging systems.

Similarly, replacing our in-house FLAME-based reconstruction with the EMOCA model could produce higher-quality 3D point clouds and potentially improve FER performance. However, it remains an open question whether the quality of the \acf{JCAS}-generated data is comparable to that produced by EMOCA, or if it exhibits considerably higher levels of noise. This distinction is critical, as optimizing FER performance in practical scenarios requires training on data that closely reflects the characteristics that the input model will encounter in deployment.

\begin{figure*}[!t]
\centering
\subfigure[Face and glasses model]{
\includegraphics[width=0.188\textwidth]{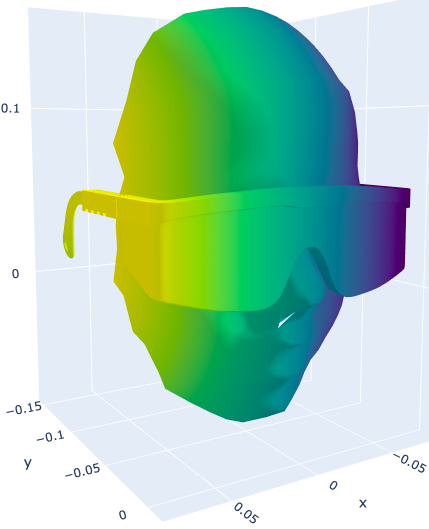}}
\subfigure[THz beamforming and beamsteering]{
\includegraphics[width=0.318\textwidth]{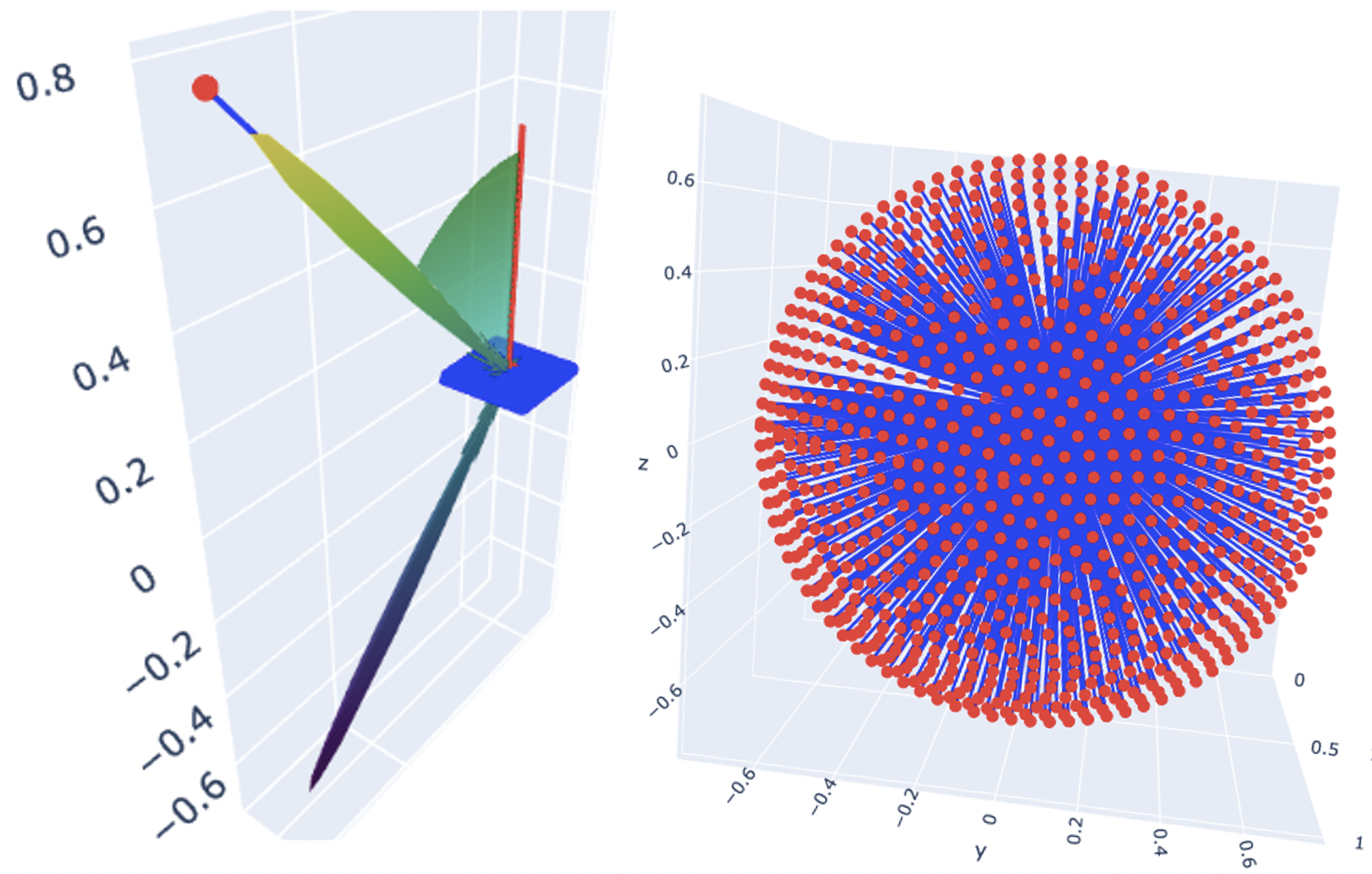}}
\subfigure[Positioning of transmitting arrays]{
\includegraphics[width=0.238\textwidth]{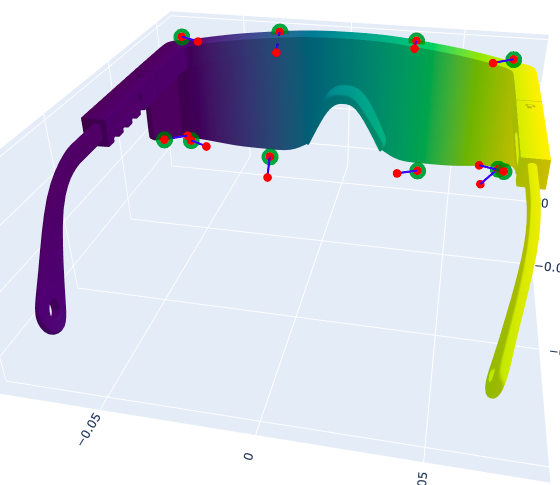}}
\subfigure[3D imaging coverage]{
\includegraphics[width=0.188\textwidth]{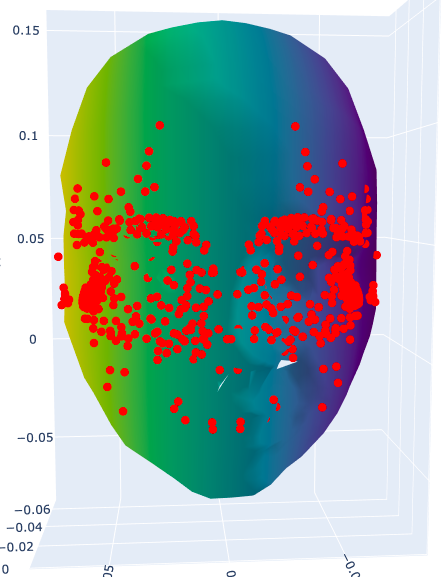}}
\caption{Illustrative example of THz array configurations and beam coverage for facial surface 3D imaging}
\vspace{-4mm}
\label{fig:arrayPattern}
\end{figure*}

\subsection{3D Imaging Using \ac{JCAS}}
Our system aims to leverage short-range \ac{JCAS}, a novel approach that presents unique challenges and opportunities. 
Designing transceivers for such a system will involve addressing the spatial constraints of wearable devices like glasses. 
Optimizing the placement of transceivers is crucial for maximizing \ac{3D} imaging coverage while maintaining the system's compactness and practicality. 

As a motivating example illustrating a foundational approach for future development, we explored the potential of using phased arrays in the \ac{THz} band for 3D facial surface recognition. 
We began by modeling the complex topography of the human face~\cite{barbarino2009development,munn2018changes}, phased array geometry~\cite{hoang2007low,tousi201414}, and short-range \ac{THz} wave propagation~\cite{piesiewicz2007short,nagatsuma2016practical}. 
Our phased arrays allowed for precise control over the direction of emitted \ac{EM} waves by adjusting the phase at each element. 
The operational range for the considered beamforming and beamsteering model is controlled within the elevation and azimuth ranges of [-50º,50º] (Figure~\ref{fig:arrayPattern}.b).
Utilizing the unique properties of \ac{THz} radiation, primarily its high spatial resolution, we aimed to achieve detailed facial surface mapping. 
The face was modeled as a 3D surface with varying topography, captured using a Gaussian map triangulation method (cf., Figure~\ref{fig:arrayPattern}.a). 
This setup allowed the THz wavefront to interact with the facial surface, reflecting back with varying phase delays, which were then captured by a set of single-antenna receivers distributed in a grid-like fashion throughout the frame of the glasses.
\textcolor{blue}{We emphasize that this section provides an illustrative feasibility-oriented sensing design sketch and does not constitute an implemented end-to-end THz imaging prototype.}

We parameterized the model by focusing on aspects such as the number of array elements, element spacing, and beamwidth.  
We used an $N \times N$ planar configuration of elements of transmitter arrays with element spacing set to $\lambda/2$ to achieve optimal interferometry and minimal grating lobes and feasible form-factors for wearables such as glasses. 
As shown in Figure~\ref{fig:arrayPattern}.b, a 32x32 array provided a precise 3D radiation pattern, with the array size optimized to maintain coherence even in mid- and far-field approximations. 
This configuration allowed us to cover the facial surface within the desired elevation and azimuth angles effectively.
We employed a uniform beam distribution strategy to ensure broad facial surface coverage, with strategic placement of transmitters and omnidirectional receivers to maximize detection even in complex topography. 
Figure~\ref{fig:arrayPattern}.b also illustrates the use of a diamond distribution to achieve near-uniform coverage in all critical areas of the face. 

Our initial configuration included a setup of 10 arrays of 32x32 transmitting elements, with their central orientation for the phased-array radiation emission as depicted in Figure~\ref{fig:arrayPattern}.c.
This setup yielded a comprehensive view of the face, as shown in Figure~\ref{fig:arrayPattern}.d. 
As visible in the figure, not only locations near glasses were imaged by the considered system, but also for example points on the forehead and near mouth. 
This is due to the phased arrays' high spatial resolution, enabling the imaging signals to be reflected back to the receivers even from small aberrations. 
Regardless of the encouraging coverage, a more thorough evaluation with an enhanced level for realism is required. 
Based on this, an optimization on the number of transmitters and receivers should be performed to further enhance the coverage of \ac{JCAS} imaging. 

The utilized PointNet++ model shows promise for deployment on edge devices like smart glasses. 
When quantized to INT8, the model footprint reduces to a quarter, enabling real-time inference within $\sim$10 ms on embedded \acp{NPU} ($\sim$1 TOPS) and within $\sim$20 ms on FP16-capable mobile \acp{GPU}, well within real-time performance bounds~\cite{paolieri2024edgesys, xie2023tinyissimoyolo}. 
\ac{THz}-based sensing used to capture the 3D facial input features \textcolor{blue}{is expected to exhibit} sub-millisecond latency, not constituting a bottleneck. 
\ac{CPU}-only execution remains feasible ($\sim$100 ms) for prototyping but is less suited for continuous wearable use.

Future work will explore additional optimization strategies to further reduce compute overhead while maintaining accuracy. 
For example, we will consider structured channel pruning (20--40\%) and hardware-aware neural architecture search, which are expected to lower the number of \acp{FLOP} and improve efficiency on resource-constrained \acp{SoC}. 
Furthermore, we will consider adopting faster spatial indexing methods for pre-processing stages or learned approximations~\cite{zhu2024splatting, zhao2025pointnetv3}.

\subsection{System-Level Considerations}

To optimize the system, targeting specific regions of the face that carry the most relevant information for \ac{FER} could be beneficial. Previous studies have shown that focusing on a limited number of characteristic points can maintain high classification accuracy for 2D RGB images~\cite{wang2022learning}. Applying a similar approach to 3D facial pointclouds could enhance system efficiency. A focused analysis of critical facial regions might reduce the amount of data processed, leading to energy savings and faster classification rates. 
Investigating this approach for 3D \ac{FER} could fill a gap in the literature and provide valuable insights into improving system performance and energy efficiency.

\begin{figure}
\centering
  \includegraphics[width=0.8\linewidth]{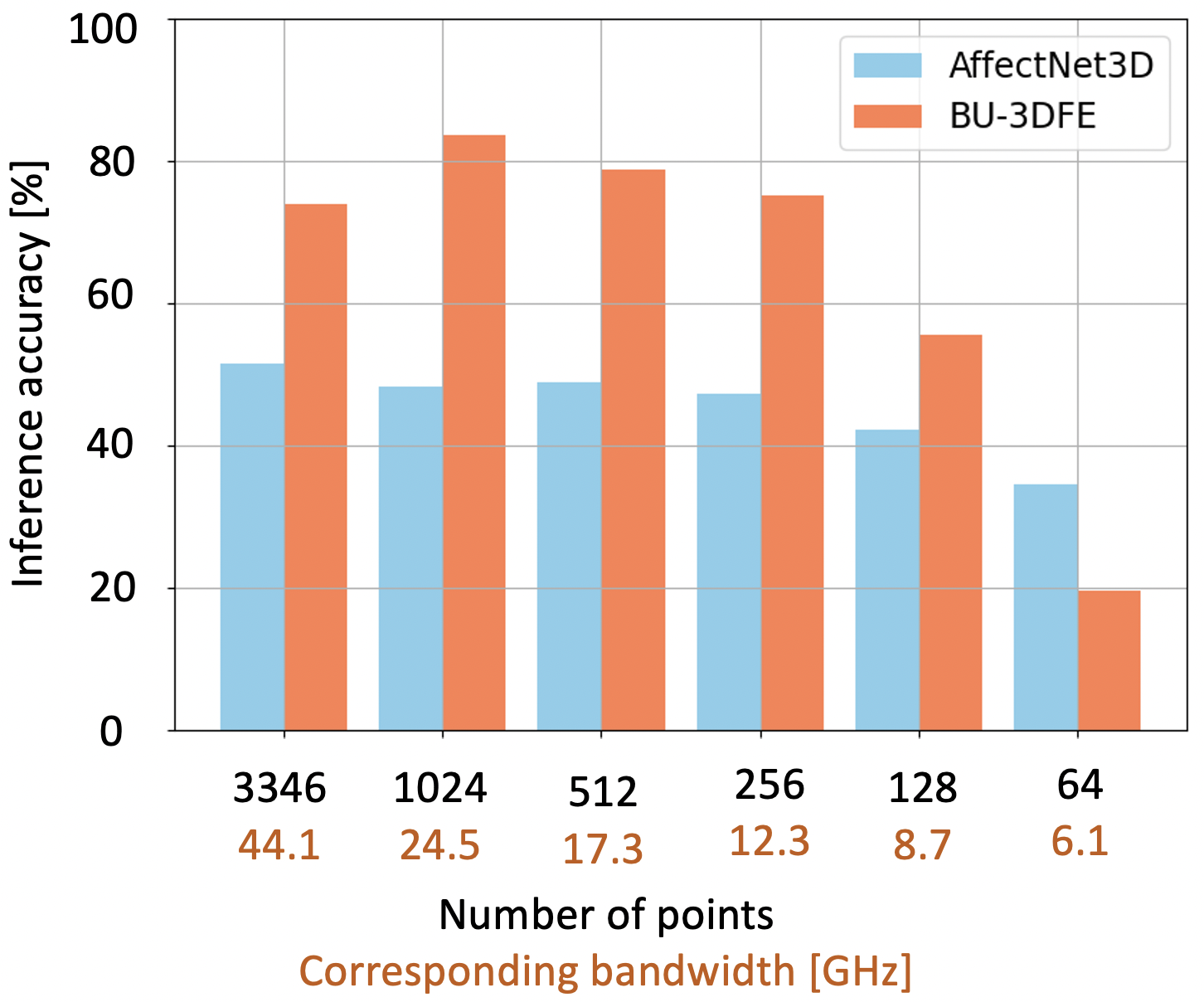}
  \caption{Classification accuracy for different number of downsampled points and corresponding bandwidths required for 3D imaging}
  \vspace{-5mm}
  \label{fig:downsampling}
\end{figure}

Additionally, determining the lowest bandwidth that maintains \ac{FER} accuracy requires investigating the impact of pointcloud downsampling and noise introduction. By analyzing the correlation between noise levels and 3D imaging resolution, we can identify the minimum bandwidth necessary to ensure effective emotion recognition.
As a simplified example, we assume the average human face has dimensions of 16$\times$24~cm, resulting in a surface area of 0.0384~m$^2$. For facial recognition using 3D pointclouds, we vary the point density as shown in Figure~\ref{fig:downsampling}, assuming homogeneous coverage over the face. 
The spacing between points, calculated as $\text{Surface Area}/{N}^{1/2}$, defines the required resolution for accurate recognition.
The radar system’s bandwidth $B$ is derived using $X_r = \frac{c}{2B}$, where $X_r$ is the required resolution. As the number of points increases, the required bandwidth rises. 
As visible from the figure \textcolor{blue}{around} 256 points, we find a ``sweet spot'' where bandwidth minimization is achieved without significant loss in FER performance.

This bandwidth is available at \ac{THz} frequencies and assuming short-range 3D imaging, offering windows exceeding 10~GHz of bandwidth~\cite{abadal2019media}. 
However, implementing transceivers that operate at these high frequencies presents significant challenges.
One of the primary challenges is the development of compact, high-performance antennas and circuits. 
At such frequencies, even small variations in the physical dimensions of components can lead to substantial performance degradation due to the shorter wavelengths involved.
Designing antennas that are small enough to be integrated into mobile devices or other compact form factors while maintaining efficiency and signal strength is a major hurdle.

Another issue is the power consumption of transceivers operating at \ac{THz} frequencies. The power requirements increase significantly as frequency increases, which can lead to heat dissipation problems and limit the practical use of such systems in battery-powered or portable devices. Developing efficient power amplifiers and low-power circuitry that can operate at such frequencies without excessive energy demands is an ongoing challenge.
Overcoming these challenges will be critical for enabling the widespread use of \ac{JCAS} systems, especially in compact and mobile applications.

\subsection{Implementational Challenges}
While our work envisions embedding \ac{THz} arrays in smart glasses for \textcolor{blue}{privacy-aware} \ac{3D} \ac{FER}, several engineering challenges are foreseen. 
Miniaturizing \ac{THz} transceivers within a wearable form factor introduces constraints in power consumption, thermal dissipation, and antenna array design. 
Developing compact, low-power \ac{THz} transceivers with efficient cooling mechanisms is a crucial aspect of making this approach practical. 
Additionally, the glass-mounted arrays must be optimized to balance field of view and depth accuracy while maintaining user comfort and aesthetics. 
Future work should explore hardware design strategies, such as integrating advanced beamforming techniques, using energy-efficient graphene-based \ac{THz} transceivers, and optimizing antenna array geometry for improved coverage.

Another open question is the quality of \ac{3D} pointclouds generated from a glasses-mounted \ac{THz} system. 
Unlike full-head imaging setups, glasses provide a limited viewpoint, potentially leading to occlusions and incomplete reconstructions. 
A thorough study on pointcloud usability is needed, including how missing regions impact FER accuracy and whether computational techniques, such as inpainting or fusing multi-view captures, can mitigate limitations. 
Future research should include simulation-based evaluations of THz imaging coverage from a wearable form factor, alongside practical experiments using scaled-down THz imaging prototypes.

Despite its potential privacy advantages, \ac{3D} FER is not completely free of risks. High-resolution scans may still allow shape-based facial re-identification, and adversarial techniques could potentially reconstruct identifiable facial features from 3D pointclouds. 
\textcolor{blue}{Accordingly, while we motivate privacy awareness through texture-free geometry and limited coverage, we do not claim formal privacy guarantees in this work.}
To mitigate these risks, further work is needed to explore differential privacy techniques, noise injection methods, and adaptive downsampling strategies to balance privacy and accuracy in 3D FER models. Additionally, hybrid FER approaches that combine 3D pointclouds with thermal imaging may offer \textcolor{blue}{complementary} protection by reducing reliance on identity-revealing geometric cues while retaining emotionally relevant physiological signals.


\section{Conclusion}
\label{sec:conclusion}
This work addresses the critical need for \textcolor{blue}{privacy-aware} data generation in \ac{FER}, particularly in light of the limitations posed by current \ac{EU} regulations on 2D image-based methods. By leveraging the \ac{JCAS} paradigm, we offer an alternative through detailed 3D facial imaging via wearable sensors.
To support the integration of \ac{JCAS} into the \ac{DL}-based \ac{FER} research community, we introduced a \ac{FLAME}-flavoured method for generating 3D \ac{FER} databases from existing 2D datasets. Our creation of a 3D version of the AffectNet database and subsequent training of the PointNet++ 3D FER \ac{DL} model, demonstrated promising results, with significant improvements in classification accuracy following the application of a face-isolating data refining pipeline.
Additionally, we showed that further fine-tuning the PointNet++ model on a limited portion of the unseen 3D FER database BU3DFE results in effective performance across the entire database.

To evaluate the system's applicability to wearable devices with limited facial coverage, such as smart glasses, we performed masking experiments on the reconstructed pointclouds, simulating the partial visibility conditions of \ac{JCAS}-enabled wearables. These experiments \textcolor{blue}{provide a practical utility indicator under constrained observability} and hint at the fact that emotion classification accuracy could be maintained even under reduced coverage, highlighting the feasibility of deploying the approach in practical continuous-monitoring scenarios.

Although our exploration of \ac{THz}-based imaging for \ac{3D} \ac{FER} remains theoretical, our proposed and implemented pipeline for generating and processing \ac{3D} facial pointclouds serves as a strong foundation for future practical developments. 
By demonstrating that \ac{3D} \ac{FER} models can be effectively trained using synthetic \ac{3D} datasets derived from \ac{2D} images, this work provides a crucial stepping stone towards real-world applications of \textcolor{blue}{privacy-aware} emotion recognition. \textcolor{blue}{We emphasize that, while texture-free geometry and reduced facial coverage can mitigate identity disclosure risk, this work does not provide a formal privacy guarantee or an empirical privacy-attack evaluation.} The proposed framework not only validates the feasibility of \ac{3D} \ac{FER} but also highlights the potential of integrating \ac{JCAS}-enabled sensing into wearable devices. Future efforts in \ac{THz}-based imaging can build upon our findings, refining the hardware and imaging pipeline to achieve high-quality \ac{3D} reconstructions suitable for real-time emotion recognition.



\renewcommand{\bibfont}{\footnotesize}
\printbibliography



\section*{Biographies}
\vspace{-35pt}

\begin{IEEEbiography}[{\includegraphics[width=1in,height=1.25in,clip,keepaspectratio]{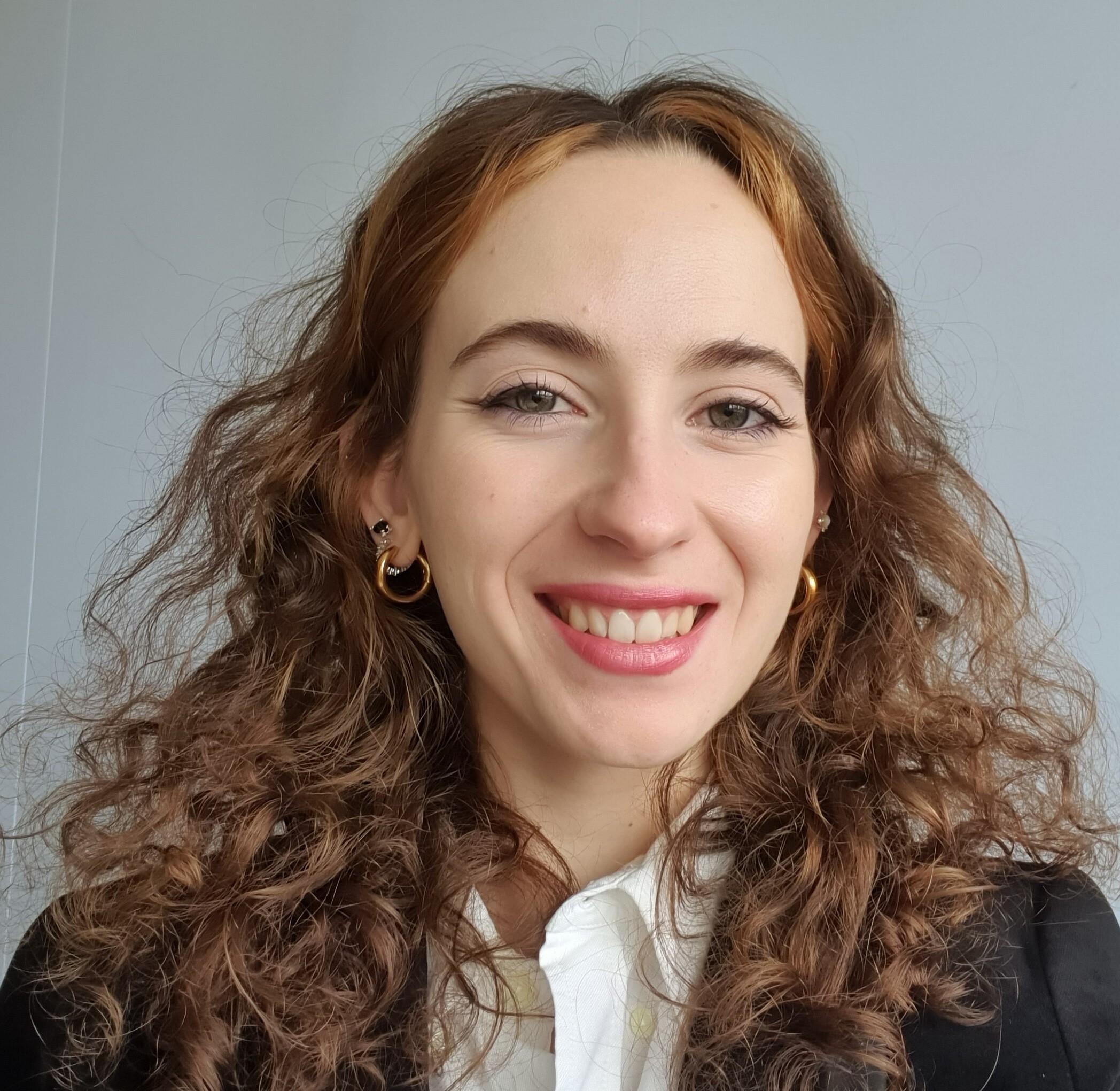}}]{Laura Rayon} is a Telecommunications engineer with MSc in Audiovisual Signal Processing and AI from UPC BarcelonaTech. Founding member of the RITA team mission, the winning project of the IEEE GRSS 2nd Student Grand Challenge, her research has included mission design and path planning algorithms for satellite networks, and a collaboration with i2CAT on Affective Computing. After working on the integration of an ethernet-based data acquisition protocol for CERN's Quench Protection System, she is currently a Rust developer.
\vspace{-20pt}
\end{IEEEbiography}

\begin{IEEEbiography}[{\includegraphics[width=1in,height=1.25in,clip,keepaspectratio]{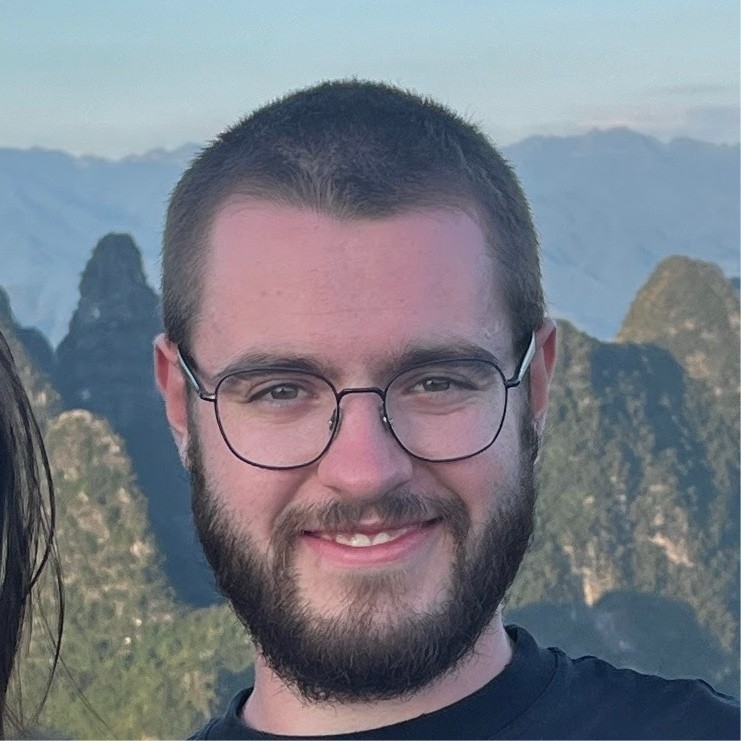}}]{Jasper De Laet} received his MSc in Artificial Intelligence and Data Science from the University of Antwerp, Belgium. He conducted research on the interpretability and explainability of spiking neural networks in collaboration with sqIRL/IDLab–imec and the University of Antwerp. His research interests include affective computing, explainable artificial intelligence, human–AI interaction, and spiking neural networks. He is currently working on the development of AI-driven applications with a focus on software and cloud-based systems.
\vspace{-20pt}
\end{IEEEbiography}

\begin{IEEEbiography}[{\includegraphics[width=1in,height=1.25in,clip,keepaspectratio]{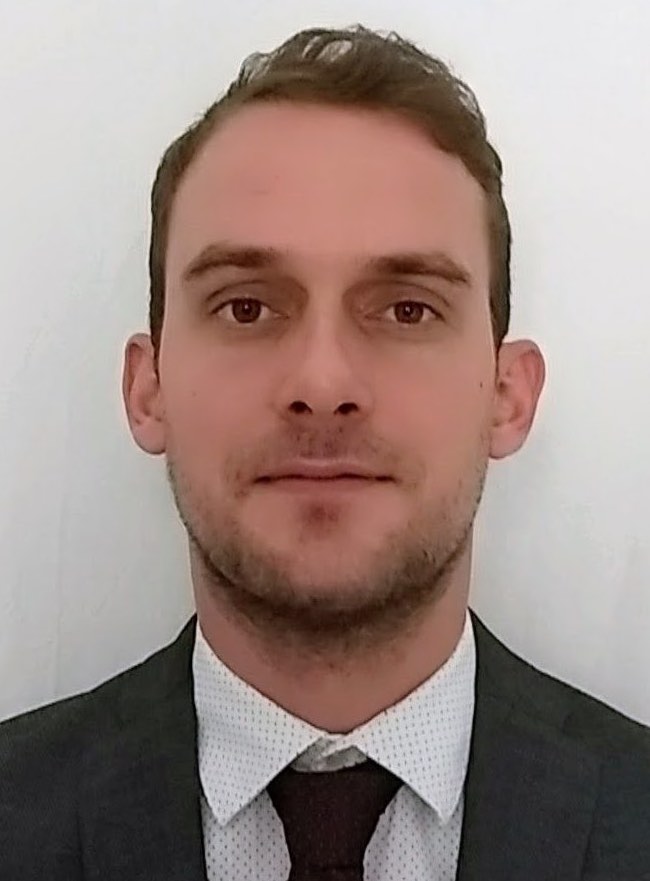}}]{Filip Lemic} is a senior researcher and a research lead at the i2Cat Foundation, and a visiting researcher at the University of Zagreb. He held positions at the University of Antwerp, imec, Universitat Politècnica de Catalunya, University of California at Berkeley, Shanghai Jiao Tong University, FIWARE Foundation, and Technische Universität Berlin. He received his M.Sc. and Ph.D. from the University of Zagreb and Technische Universität Berlin, respectively. 
\vspace{-20pt}
\end{IEEEbiography}

\begin{IEEEbiography}[{\includegraphics[width=1in,height=1.25in,clip,keepaspectratio]{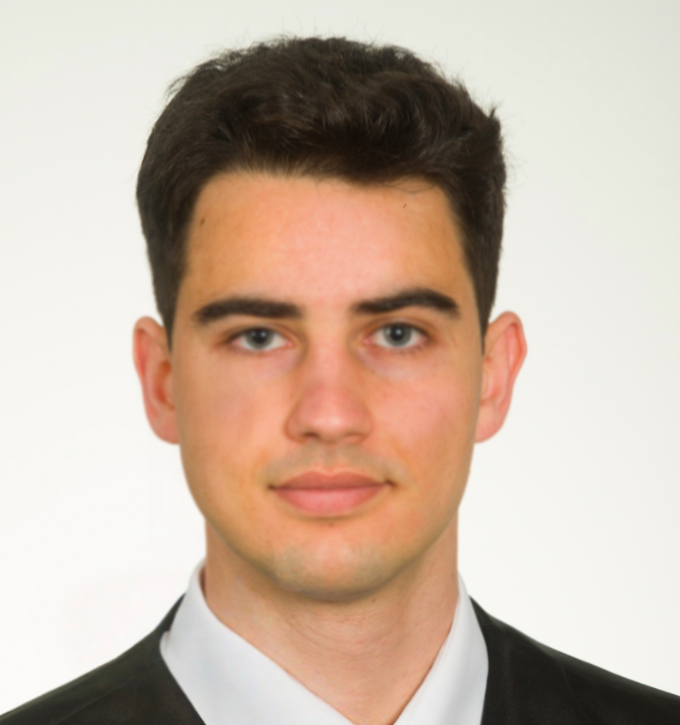}}]{Pau Sabater} is a junior researcher at the Ultrabroadband Nanonetworking Laboratory, Boston, MA, within a visiting student program at Department of Electrical and Computer Engineering, Northeastern University, where his work focuses on THz-based wavefront engineering for long-range wireless communications. He is an undergraduate student in Mathematics (B.Sc.) and Physics (B.E.) at UPC BarcelonaTech, Spain, under the CFIS excellence program. He has conducted research at i2CAT Foundation, the Institute of Robotics and Industrial Informatics under a CSIC-UPC Research Grant, and Esperanto Technologies. He also serves as a Data Scientist at REVER (YC-S22).
\vspace{-20pt}
\end{IEEEbiography}

\begin{IEEEbiography}[{\includegraphics[width=1in,height=1.25in,clip,keepaspectratio]{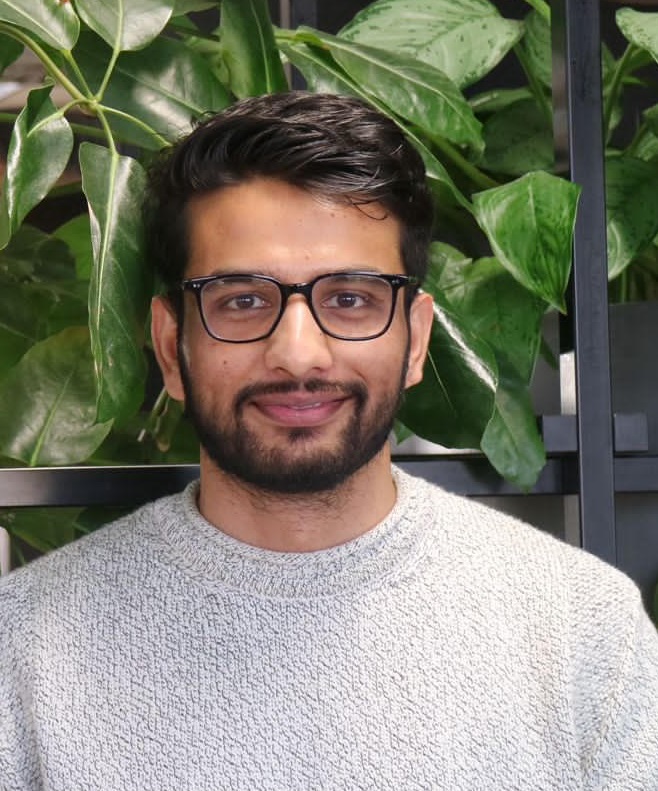}}]{Nabeel Nisar Bhat} 
is a Ph.D. researcher in the field of Joint Communication and Sensing at the IDLab research group (University of Antwerp) and imec research institute, Belgium. He obtained his M.Sc. (2021) in Communications and Computer Networks Engineering at Politecnico di Torino. His current research focuses on leveraging mmWave communication signals for pose estimation in Extended Reality applications. He has experience in signal processing, wireless communications, and deep learning.
\vspace{-20pt}
\end{IEEEbiography}

\begin{IEEEbiography}[{\includegraphics[width=1in,height=1.25in,clip,keepaspectratio]{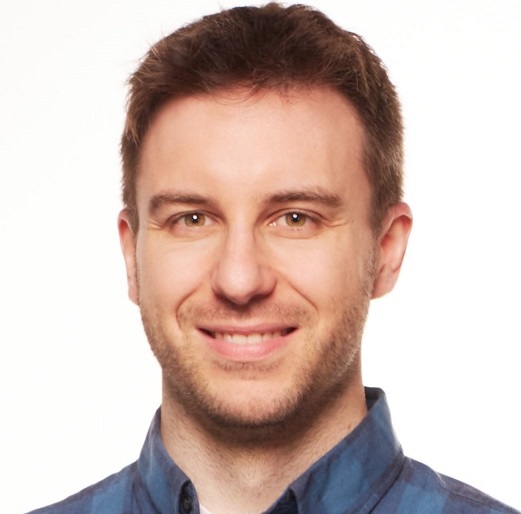}}]{Sergi Abadal} 
is an associate professor at the Universitat Politècnica de Catalunya (UPC) and the recipient of a Starting Grant from the European Research Council (ERC). He holds editorial positions in journals such as the IEEE TMC, IEEE TCAD, or IEEE JETCAS. He has served as a TPC member of more than 40 conferences and has published over 150 articles in top-tier journals and conferences. His current research interests are in the areas of wireless communications in extreme environments and its applications.
\vspace{-20pt}
\end{IEEEbiography}

\begin{IEEEbiography}[{\includegraphics[width=1in,height=1.25in,clip,keepaspectratio]{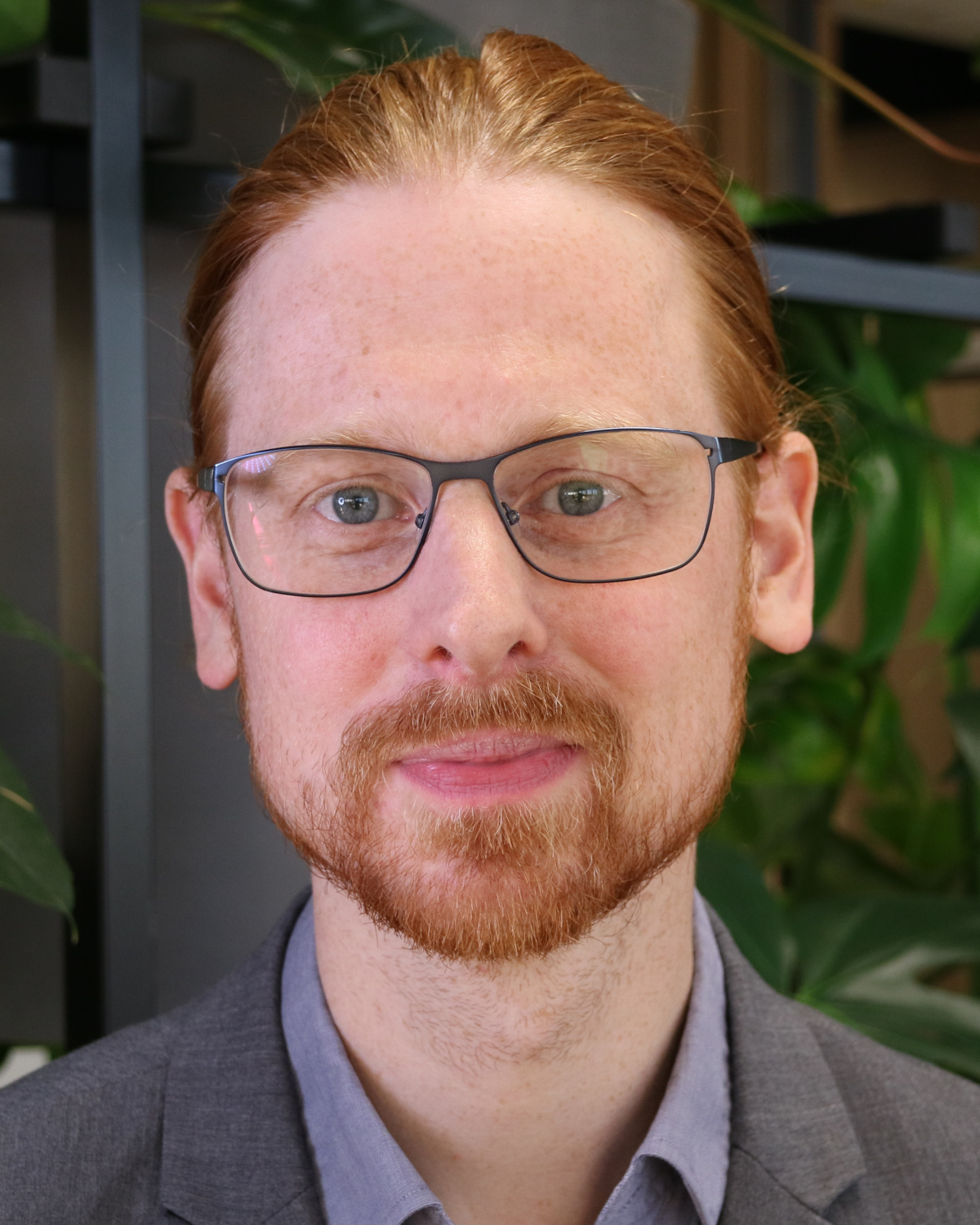}}]{Jeroen Famaey}
is an associate professor at the University of Antwerp, Belgium, and a senior researcher at IMEC, Belgium. He leads the Perceptive Radio Systems lab at the IDLab research group, performing research on wireless communications and sensing. His current research interests include low-power distributed machine learning and wireless communications for Ambient IoT, as well as data-driven integrated sensing and communications. He has co-authored over 200 articles, published in international peer-reviewed journals and conference proceedings.
\vspace{-20pt}
\end{IEEEbiography}

\begin{IEEEbiography}[{\includegraphics[width=1in,height=1.25in,clip,keepaspectratio]{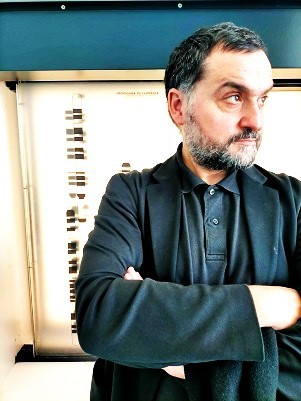}}]{Eduard Alarc\'on} 
received the M.Sc. and Ph.D. degrees in EE from UPC BarcelonaTech, Spain, in 1995 and 2000, respectively. He is a Full Professor with the School of Telecommunications, UPC. He was a Visiting Professor with the University of Colorado at Boulder, USA, and the Royal Institute of Technology (KTH), Stockholm. He has been involved in different European (H2020 FET-Open and FlagERA) and U.S. (DARPA and NSF) research and development projects within his research interests including the areas of on chip energy management, wireless networks-on-chip, machine learning accelerator architectures, nanotechnology-enabled wireless communications and quantum compute architectures. He has received major research awards and fellowships, along with a national master’s study award, and has held senior IEEE Circuits and Systems leadership roles. He has also led and organized multiple flagship conferences and special sessions.
\vspace{-20pt}
\end{IEEEbiography}

\begin{IEEEbiography}[{\includegraphics[width=1in,height=1.25in,clip,keepaspectratio]{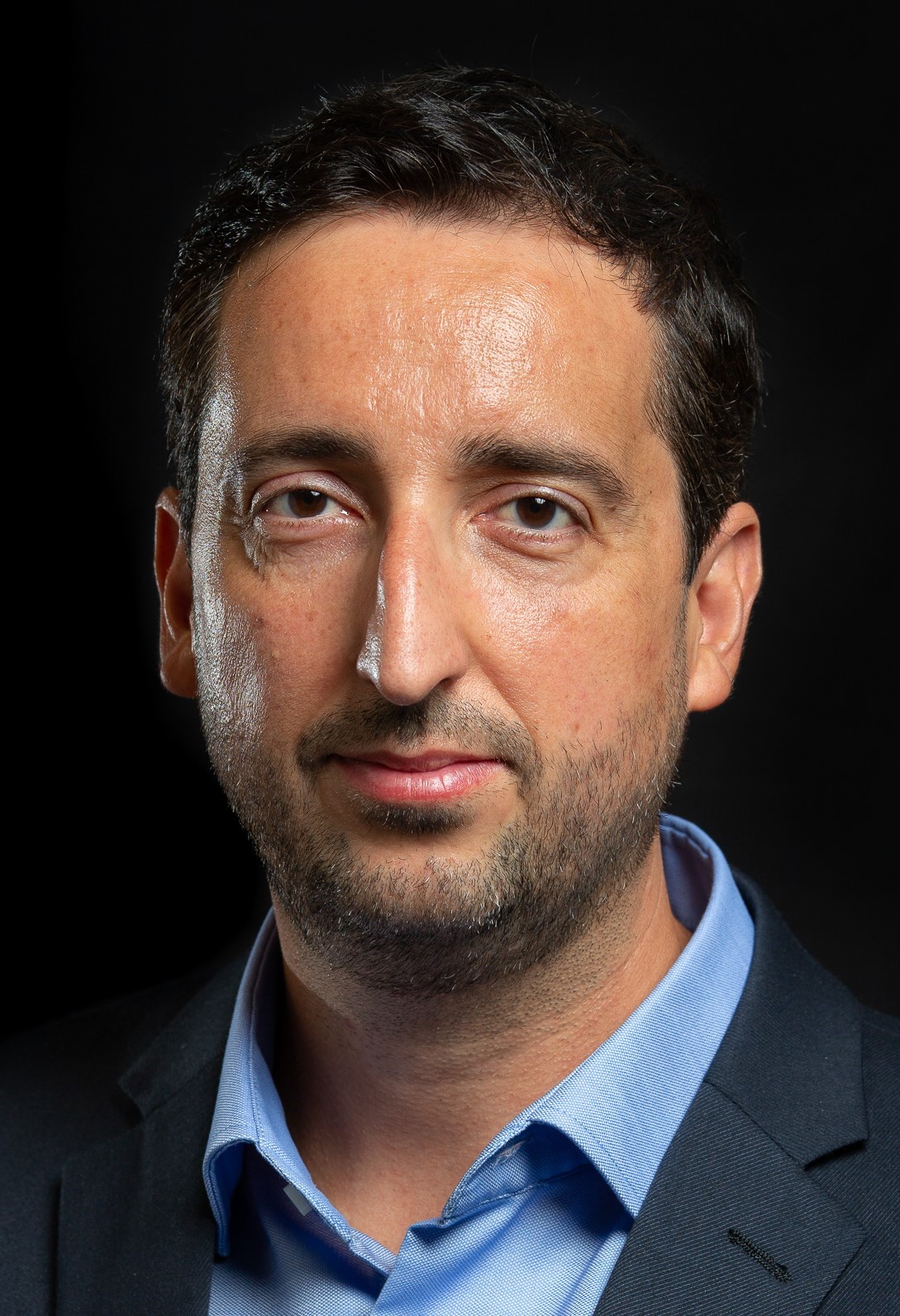}}]{Xavier Costa-Pérez} is an ICREA Research Professor and Scientific Director at the i2cat Research Center, concurrently leading 5G/6G R\&D at NEC Laboratories Europe. His team consistently delivers impactful research, evidenced by publications in top-tier scientific venues and numerous awards for successful technology transfers. Notably, his innovations have been integrated into commercial mobile phones, base stations, and network management systems, and have spurred the creation of multiple start-ups. He has also served on organizing committees for prominent conferences (e.g., ACM MOBICOM, IEEE INFOCOM) and as an Editor for leading journals like IEEE Transactions on Mobile Computing (TMC), IEEE Transactions on Communications (TCOM), as well as Elsevier Computer Communications (COMCOM). Dr. Costa-Pérez holds M.Sc. and Ph.D. degrees in Telecommunications from the Polytechnic University of Catalonia (UPC), receiving a national award for his doctoral thesis. He is the inventor of approximately 100 granted patents.
\vspace{-10pt}
\end{IEEEbiography}

\end{document}